\documentclass[10pt,twocolumn,letterpaper]{article}

\usepackage{cvpr}        
\usepackage[pagebackref,breaklinks,colorlinks,citecolor=cvprblue]{hyperref}
\usepackage{amssymb}
\usepackage{mathrsfs}
\usepackage{amsmath}
\usepackage{booktabs}               
\usepackage{nicematrix}
\usepackage{graphicx}
\usepackage{wrapfig}
\usepackage{arydshln}
\usepackage{pifont}
\usepackage{caption}
\usepackage{subcaption}
\usepackage{tablefootnote}    
\usepackage{graphicx}
\usepackage{multirow} 
\usepackage{bm}
\newtheorem{theorem}{Theorem}[section]
\usepackage{algorithm}
\usepackage{algpseudocode}

\definecolor{cvprblue}{rgb}{0.21,0.49,0.74}

\def\confName{CVPR}
\def\confYear{2024}
\def\ourmethod{EAGLE}

\makeatletter
\newcommand{\myfnsymbol}[1]{%
  \expandafter\@myfnsymbol\csname c@#1\endcsname
}
\newcommand{\@myfnsymbol}[1]{%
  \ifcase #1
  \or 1
  \or 2
  \or 3
  \or \TextOrMath{\textasteriskcentered}{*}
  \or \TextOrMath{\textdagger}{\dagger}
  \fi
}
\newcommand{\affiliationA}{\@myfnsymbol{1}}
\newcommand{\affiliationB}{\@myfnsymbol{2}}
\newcommand{\affiliationC}{\@myfnsymbol{3}}
\newcommand{\equalcontributor}{\@myfnsymbol{4}}
\newcommand{\correspondingA}{\@myfnsymbol{5}}
\makeatother

\title{Efficient Training of Generalizable Visuomotor Policies via\\ Control-Aware Augmentation}

\author{
  Yinuo Zhao\textsuperscript{\affiliationA, \equalcontributor},
  Kun Wu\textsuperscript{\affiliationB, \equalcontributor},
  Tianjiao Yi\textsuperscript{\affiliationA},
  Zhiyuan Xu\textsuperscript{\affiliationB},
  Xiaozhu Ju\textsuperscript{\affiliationB},
  Zhengping Che\textsuperscript{\affiliationB}\\
  Chi Harold Liu\textsuperscript{\affiliationA},
  Jian Tang\textsuperscript{\affiliationB}
}

\begin{document}
\renewcommand{\thefootnote}{\myfnsymbol{footnote}}
\maketitle
\footnotetext[1]{Beijing Institute of Technology, \{ynzhao, tjyi\}@bit.edu.cn, liuchi02@gmail.com}%
\footnotetext[2]{Beijing Innovation Center of Humanoid Robotics, \{gongda.wu, eric.xu, jason.ju, z.che, jian.tang\}@x-humanoid.com}%
\footnotetext[4]{These authors contributed equally.}%
\setcounter{footnote}{0}
\renewcommand{\thefootnote}{\fnsymbol{footnote}}

\begin{abstract}
Improving generalization is one key challenge in embodied AI, where obtaining large-scale datasets across diverse scenarios is costly.
Traditional weak augmentations, such as cropping and flipping, are insufficient for improving a model's performance in new environments.
Existing data augmentation methods often disrupt task-relevant information in images, potentially degrading performance.
To overcome these challenges, we introduce \textbf{EAGLE}—an \textbf{E}fficient tr\textbf{A}ining framework for \textbf{G}enera\textbf{L}izabl\textbf{E} visuomotor policies—that improves upon existing methods by: 1) enhancing generalization by applying augmentation only to control-related regions identified through a self-supervised control-aware mask; and 2) improving training stability and efficiency by distilling knowledge from an expert to a visuomotor student policy,  
which is then deployed to unseen environments without further fine-tuning. Comprehensive experiments on three domains—including the DMControl Generalization Benchmark, the enhanced Robot Manipulation Distraction Benchmark, and a long-sequential drawer-opening task—validate the effectiveness of  our method. Project website:\href{https://vrl-eagle.github.io/}{https://vrl-eagle.github.io/}
\end{abstract} 

\section{Introduction}
End-to-end visuomotor policies learn low-level controls directly from high-dimensional inputs using either reinforcement learning or imitation learning algorithms. These learning paradigms yield promising results in embodied AI tasks such as robot manipulation~\cite{chen2020adversarial, kalashnikov2018scalable}, autonomous navigation~\cite{banino2018vector}, and locomotion~\cite{yarats2021mastering, yu2021visual}. However, visuomotor policies heavily rely on visual inputs for decision-making and control, making them susceptible to performance degradation when faced with changes in background, distractors, or viewpoints. This deficiency cannot be mitigated through reinforcement nor imitation learning alone.

One promising technique to reduce the impact of these visual discrepancies is Data Augmentation (DA)~\citep{yarats2021image,yarats2021mastering,hansen2021stabilizing,schwarzer2020data,fan2021secant}. \textit{Weak augmentations}, like random cropping, rotation, and flipping, consistently enhance generalization, but with modest improvements. In contrast, \textit{strong augmentations} such as \textit{random conv}~\citep{lee2019network} and \textit{random overlay}~\citep{hansen2021generalization}, boosting generalization capabilities through significantly diversifying the data. Followed by this, algorithms~\citep{hansen2021stabilizing, fan2021secant, choi2023environment} propose to maintain consistency in visual features or control values between augmented and original observations. Nevertheless, \textit{strong augmentations} can indiscriminately distort the entire observation space, disrupting the control-related environmental structures and dynamics captured in the data. This often complicates training and destabilizes both learning and testing phases. While previous research has focused on augmenting specific areas within the observational space, these methods typically depend on dense reward functions \citep{fu2021learning, bertoin2022look} and are limited to identifying dynamic objects \citep{wang2021unsupervised} without considering their task relevance. Recently, vision foundation models like SAM~\citep{kirillov2023segment} have shown strong generalization abilities. However, they still require fine-tuning or human-given priors to identify task-relevant regions in the observation space. Therefore, automatically identifying control-related pixels for generalizable visuomotor policies still remains challenging, especially in tasks with long horizons and non-informative rewards.

To enhance generalization and maintain stability, in this paper, we propose an \textbf{E}fficient tr\textbf{A}ining framework for \textbf{G}enera\textbf{L}izabl\textbf{E} visuomotor policies (\ourmethod) that can identify control-related information and facilitate zero-shot deployment in unseen environments. Instead of learning a generalizable visuomotor policy directly, we adopt a distillation approach for efficient and stable training. Specifically, \ourmethod~consists of two jointly optimized modules:
1) A control-aware augmentation module that adaptively learns to highlight control-related pixels using spatial-temporal data from the following distillation module, employing a self-supervised reconstruction task;
2) A privilege-guided distillation module to extract control knowledge from an expert agent trained with a standard deep reinforcement learning (DRL) algorithm. The DRL expert receives low-level environment states. It is worth noting that this privileged information is readily obtainable~\citep{pinto2017asymmetric,chen2020learning,lee2020learning,salter2021attention,galashov2022data} in a robotic simulated environment, and would only be accessed during training.

To evaluate the zero-shot generalization abilities, we conducted experiments over three domains, including the commonly used DMControl Generalization Benchmark~\citep{hansen2021generalization}, an enhanced Robot Manipulation Distraction Benchmark based on~\citep{jangir2022look} with visual changes on the control-irrelevant background and distractors, a self-built challenging drawer-operation generalization platform based on NVIDIA Isaac Sim~\citep{makoviychuk2isaac}. 
Extensive experimental results show that our method advances the state-of-the-art in terms of generalization ability and computation cost.

We summarize our contributions as follows:
\begin{itemize}
    \item We propose an efficient training framework for generalizable visuomotor policies to achieve zero-shot generalization to unseen environments with visual changes.
    \item We introduce a method to derive a control-aware mask within a self-supervised reconstruction structure, eliminating the need for additional labels or reward signals. 
    \item To enhance training stability and efficiency, we develop a strategy to learn visuomotor policies by distilling knowledge from a privileged expert pre-trained on low-level environmental states. 
    \item Extensive comparative and ablation studies across three benchmarks well validate the effectiveness of our method.     
\end{itemize}

\begin{figure*}[thp]
    \centering
    \includegraphics[width=1.0\textwidth]{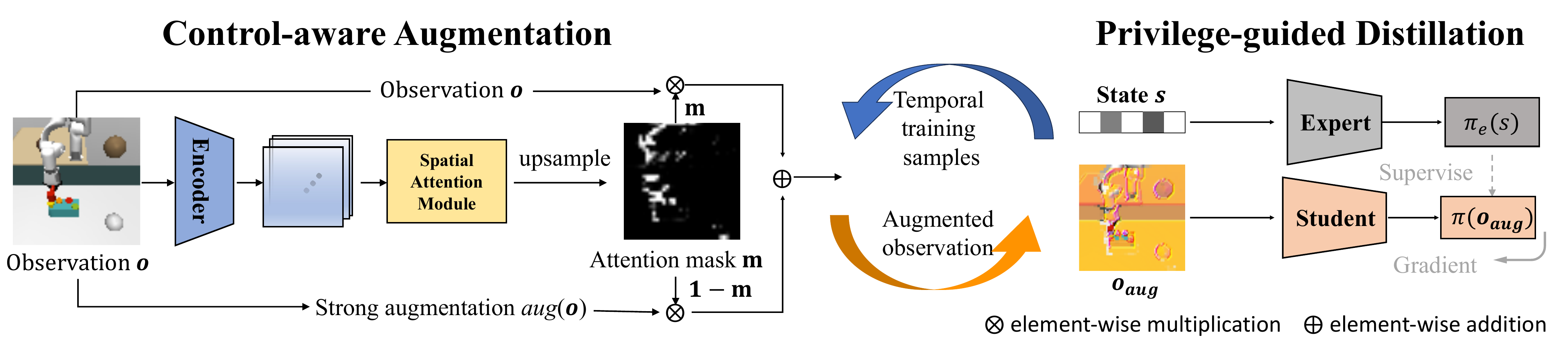}
    \caption{Overview of our method.}
    \label{fig_overall}
\end{figure*}

\section{Related works}
Generalization performance degradation results from data distribution shifts from the training environments to testing ones. Common distribution shifts include changes in visual appearance~\cite{fan2021secant, bertoin2022look, yuan2022don, li2024normalization}, object distractors~\cite{liu2023robust, wang2021unsupervised}, environmental structures~\cite{cobbe2019quantifying, liuunlock}, and viewpoints~\cite{yang2024movie}. In this work, we primarily focus on visual changes but also experiment with distractors and viewpoint variations.

Improving the generalization of visuomotor policies remains a significant challenge~\citep{rajeswaran2017towards, packer2018assessing, zhang2018dissection, justesen2018illuminating, machado2018revisiting, cobbe2019quantifying, wang2019generalization, cobbe2020leveraging, yarats2021improving, raileanu2020ride}. Regularization techniques like dropout~\citep{srivastava2014dropout}, entropy regularization~\citep{zhang2018study, haarnoja2018soft}, and batch normalization~\citep{igl2019generalization} are commonly used for generalization improvements in supervised learning. However, for more complex visuomotor policies, they offer limited improvements and may reduce sampling efficiency~\citep{cobbe2019quantifying, yarats2021image}.

\textbf{Data augmentation} (DA) is a simple yet efficient technique for enhancing the zero-shot generalization capabilities of visuomotor policies. \textit{Weak augmentations}, such as geometric (e.g., random cropping) and photometric (e.g., grayscale conversion) transformations, are easily integrated with various policy learning algorithms but offer limited generalization benefits. In contrast, \textit{strong augmentations}~\citep{fan2021secant,yuan2022don,bertoin2022look} introduce highly diverse samples, potentially bridging generalization gaps; however, they can lead to a diverged action distribution during training due to visual variability~\citep{yuan2022don}. DA methods generally focus on consistency between the augmented and original inputs, either in visual features—such as in SGQN~\citep{bertoin2022look}, SODA~\citep{hansen2021generalization}, and VAI~\citep{wang2021unsupervised}—or in output actions, as with SECANT~\citep{fan2021secant}, TLDA~\citep{yuan2022don}, and SVEA~\citep{hansen2021stabilizing}. \ourmethod~improves visuomotor policies by aligning their outputs directly with the expert’s, falling in the latter category. The control-aware augmentation module in our method is similar to that in VAI~\citep{wang2021unsupervised}, but focuses on identifying control-related regions rather than merely the dynamic foreground. This approach allows for the augmentation of whole control-irrelevant regions, leading to improved generalization. Besides DA, other approaches focus on auxiliary tasks or redesigning training objectives to learn invariant and robust representations, such as the information bottleneck~\citep{fan2022dribo}, bisimulation metrics\citep{zhang2020learning, kemertas2021towards}, and pretrained backbones~\citep{yuan2022pretrained}. \ourmethod~is orthogonal to these techniques and can be combined for further improvements.

\textbf{Policy distillation} has been widely explored in various scenarios and tasks~\citep{czarnecki2019distilling,chen2020learning,lee2020learning}. Similar to SECANT~\citep{fan2021secant}, our method uses distillation to maintain stability, while leveraging a privileged expert to reduce computational costs and enhance performance. Unlike SECANT~\citep{fan2021secant}, which relies on mix of augmentations, our approach shows that one well-chosen \textit{strong augmentation} method is enough for improving generalization with a more efficient training process for robot learning. In other domains, LBC~\citep{chen2020learning} trains a student policy with limited sensor input under the supervision of an expert with privileged information in the CARLA simulator~\citep{dosovitskiy2017carla} for autonomous driving tasks. Zhou \textit{et al.}~\citep{zhou2020domain} have developed a task distillation framework to transfer navigation policies between different simulators. Similarly, ITER~\citep{igl2020impact} transfers accumulated knowledge into newly initialized networks to mitigate non-stationary effects. The privileged expert, trained in simulators, is readily accessible. With the recent advances in vision foundation models~\citep{kirillov2023segment,ravi2024sam}, obtaining state information such as object positions in real environments has become easier, enabling the deployment of privileged experts in actual platforms.

\section{Method}
In this section, we introduce an efficient training framework for generalizable visuomotor policies (\ourmethod). The overall goal of \ourmethod~ is to directly learn visuomotor policies that are invariant and capable of zero-shot generalization. As shown in Fig~\ref{fig_overall}, \ourmethod~ consists of two simultaneously optimized modules: a control-aware augmentation module (left) and a privilege-guided distillation module (right). The former module retrieves temporal data from the replay buffer and conducts a self-supervised reconstruction task, accompanied by three auxiliary losses, to identify control-related pixels. The latter module augments the observation input and distills knowledge from a pretrained DRL expert (which processes only environment states) into the visuomotor student network (which processes only image observations). After training is completed, the visuomotor policy can be reliable deployed in complex environments with visual variations, without the need for fine-tuning or additional supervision.

\subsection{Control-aware Data Augmentation}
The control-aware data augmentation module is designed to learn an attention mask that efficiently identifies control-related regions, distinguishing them from irrelevant areas such as static backgrounds and unrelated objects. In this way, we can confidently use \textit{strong augmentation} without fear of compromising essential information for policy learning, thereby enhancing generalization ability. To achieve this,  we develop a lightweight mask-generating model using Convolutional Block Attention~\citep{vaswani2017attention} (CBA) and train a clear attention mask in a self-supervised manner.

\begin{figure}[t]
    \centering
    \includegraphics[width=1.0\columnwidth]{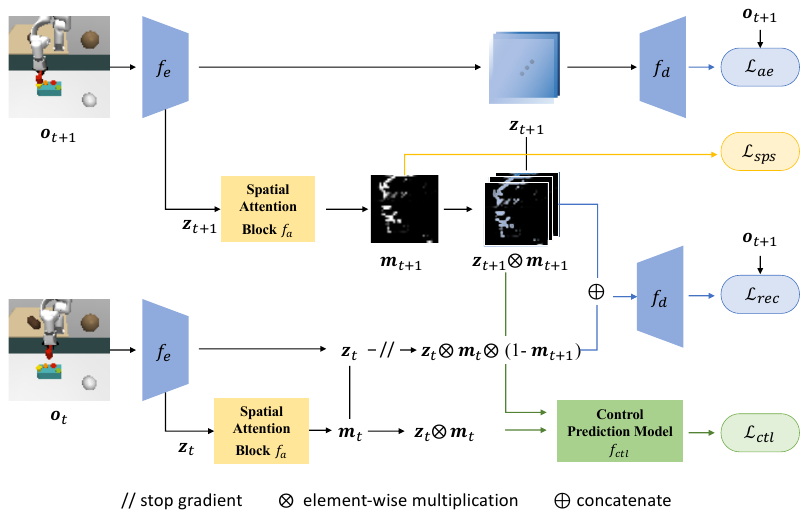}
    \caption{Control-aware data augmentation module.}
    \label{fig_att_model}
\end{figure}

As shown in Fig.~\ref{fig_att_model}, the control-aware data augmentation module consists of four networks:
1) An image encoder $f_{e}(\cdot)$ that takes the observation $\bm{o}$ as input and outputs a feature map $\bm{z}$,
2) A spatial attention block $f_{a}(\cdot)$ that receives $\bm{z}$ and outputs the attention mask $\bm{m}$,
3) A decoder $f_{d}(\cdot)$ that processes the feature maps to produce the reconstructed image $\hat{\bm{o}}$,
4) A control prediction module $f_{ctl}(\cdot)$ that utilizes control-related feature maps to generate the predicted control $\hat{a}$.

Below, we detail the methods for learning the control-aware mask that differentiates between control-related and control-irrelevant regions, introducing a self-supervised reconstruction task complemented by three auxiliary losses to facilitate efficient learning.

At the beginning, we randomly sample a source image $\bm{o}_t$ from the replay buffer and obtain its next transition $\bm{o}_{t+1}$ as the target image. We derive the feature map $\bm{z}_t=f_{e}(\bm{o}_t)$ and the attention mask $\bm{m}_t = f_{a}(\bm{z}_t)$ from the source image. Similarly, we obtain $\bm{z}_{t+1}$ and $\bm{m}_{t+1}$ from the target image. Thus, $\bm{z}_{t} \otimes \bm{m}_{t}$ and $\bm{z}_{t+1} \otimes \bm{m}_{t+1}$ specifically represent the control-related features extracted by the control-aware masks from the source and target images. Following the approach of~\citep{wang2021unsupervised}, we synthesize the latent feature $\hat{\bm{z}}_{t+1}$ for reconstructing the target image using the equation:
\begin{equation}
\label{equ: m_j}
    \hat{\bm{z}}_{t+1} = \bm{z}_{t+1} \otimes \bm{m}_{t+1} + \bm{z}_t \otimes (1 - \bm{m}_t) \otimes (1 - \bm{m}_{t+1}),
\end{equation}
where the operator $\otimes$ denotes element-wise multiplication across channels.  By multiplying the $1 - \bm{m}_{t+1}$ in the second term, we can reduce the interference from the intersection part of $1 - \bm{m}_t$ and $\bm{m}_{t+1}$, enhancing the accuracy of the latent feature for the target image. Finally, we synthesize the target image by processing $\hat{\bm{z}}_{t+1}$ through the decoder $f_{d}(\cdot)$. The reconstruction loss is calculated as follows:
\begin{equation}
    \mathcal L_{rec}(\bm{o}_t, \bm{o}_{t+1}) = \Vert f_{d}(\hat{\bm{z}}_{t+1}) - \bm{o}_{t+1} \Vert_2^2.
\end{equation}

Then, we introduce three auxiliary losses to facilitate learning a clear control-aware mask. First, we add an auto-encoder loss $\mathcal{L}_{ae}$ by directly reconstructing from the target feature $\bm{z}_{t+1}$ to capture more accurate latent information. The auto-encoder loss  is computed as follows:
\begin{equation}
 \mathcal L_{ae}(\bm{o}_{t+1})= \Vert f_{d}(\bm{z}_{t+1}) - \bm{o}_{t+1} \Vert_2^2.    
\end{equation}

Second, to extract essential control-related regions, we construct a control prediction loss $\mathcal{L}_{ctl}$ to predict the control from the control-related features of the source and target images:
\begin{equation}
    \mathcal L_{ctl}(\bm{o}_t, \bm{o}_{t+1})= \Vert f_{ctl}(\bm{z}_t \otimes \bm{m}_t, \bm{z}_{t+1} \otimes \bm{m}_{t+1}) - a_t\Vert_2^2,
\end{equation}
where $\bm{z}_t$ and $\bm{m}_t$ are the feature map and the attention mask from the source image, respectively. To ensure stable training, the gradient to $\bm{z}_t$ is detached, as shown in Fig.~\ref{fig_att_model}.

Lastly, we add a sparsity penalty loss $\mathcal{L}_{sps}$ to flexibly control the generated attention mask. $\mathcal{L}_{sps} = \Vert m_j \Vert_1.$ The overall loss function is defined as:
\begin{equation}
\label{equ:att_loss}
    \mathcal L_{att} = \mathcal L_{rec} + \mathcal L_{ae} + \beta \mathcal L_{ctl} + \lambda \mathcal{L}_{sps},
\end{equation}
where $\beta$ and $\lambda$ are hyper-parameters that balance the contribution of auxiliary losses relative to the reconstruction loss. 

The image encoder $f_{e}(\cdot)$, spatial attention block $f_{a}(\cdot)$, decoder $f_{d} (\cdot)$, and control prediction model $f_{ctl}$ are optimized simultaneously. To stabilize training, we stop gradient propagation along the branch of the source image $o_t$, as illustrated in Fig~\ref{fig_att_model}.
To apply observation-level control-aware attention masks $\bm{m}$, we directly upsample them to the observation scale and only augment control-irrelevant regions. The augmented image $\bm{o}_{aug}$ is obtained as follows:
\begin{equation}
\label{equ:aug}
    \bm{o}_{aug} = \bm{o} \otimes \bm{m} + aug(\bm{o}) \otimes (1 - \bm{m}),
\end{equation}
where $aug(\cdot)$ refers to \textit{strong augmentation} operations like \textit{random conv}~\citep{lee2019network} and \textit{random overlay}~\citep{hansen2021generalization}. This strategy allows the augmentation of control-irrelevant pixels while preserving control-related pixels for decision-making.

\subsection{Privilege-guided Policy Distillation}
While \textit{strong augmentation} introduces diverse inductive biases to reduce the generalization gaps, it also complicates policy optimization significantly~\citep{fan2021secant}. Specifically, the agent must learn to output the same action for two completely different augmented observations derived from the original observation. This requirement often leads to high variance in value function estimates, making the training process unstable for any standard RL algorithm. Therefore, instead of directly training visuomotor policies with DRL, we opt to distill knowledge from a privileged DRL agent to the student visuomotor policies.

The privilege expert $\pi_e$ receives  environmental states $\bm{s}$ as inputs. Note that using privileged control-related information is reasonable in training environments, typically controlled simulators where such information is readily available~\citep{chen2020learning,lee2020learning}.
For example, in LBC~\citep{chen2020learning}, privileged information like object positions, road features, and traffic signals is distilled into a sensorimotor agent for autonomous driving tasks. Moreover, with the rapid development of vision foundation models~\citep{kirillov2023segment,ravi2024sam}, extracting state information such as object positions in real environments has become more feasible, facilitating the use of privileged experts in actual applications.

The expert policy is trained in a standard RL paradigm and  is compatible with any RL algorithm. In practice, we build our implementation on the top of DrQv2~\citep{yarats2021mastering} due to its broad adoption in continuous control tasks~\citep{wang2022vrl3,yuanpre,haldar2023watch}.

The distillation process employs the L-2 loss, defined as:
\begin{equation}\label{equ:cloning_loss}
    \hat{\mathcal{L}}(\pi_{\theta}) = \mathbb{E}_{(\bm{o}, \bm{s}) \sim \mathcal{D}} \left[\Vert (\pi_{\theta}(o_{aug}) - \pi_e(s))\Vert_2^2\right] .
\end{equation}
Following the method in SECANT~\citep{fan2021secant}, we alternately use $\pi_e$ and $\pi$ to collect environmental samples, storing them in a replay buffer $\mathcal{D}$. At each iteration, we simultaneously
1) update the context-aware data augmentation module using Eqn.~(\ref{equ:att_loss}), and 2) optimize the visuomotor policy according to Eqn.~(\ref{equ:cloning_loss}).

\subsection{Theoretical Insights}
In ~\citep{hernandez2018data}, the authors study the effect of data augmentation in the case of binary classification under Rademacher complexity. Here, we also provide theoretical Insights behind \ourmethod~under Rademacher complexity. 

The visuomotor policy network is parameterized by $\theta$ and outputs a $k$-dimensional continuous action $\pi_{\theta}(\bm{o}) \in \mathbb{R}^k$. We aim to compute the Rademacher complexity of the policy function class: $\mathcal{F} = \left\{ \pi_{\theta} \mid \theta \in \Theta \right\}$. 
For each dimension $j$, the empirical Rademacher complexity is given by:
\begin{equation}
    \hat{\mathfrak{R}}_N(\mathcal{F}_j) = \frac{1}{N} \, \mathbb{E}_{\boldsymbol{\sigma}} \left[ \sup_{\theta \in \Theta} \sum_{i=1}^N \sigma_i \, \pi_{\theta, j}(\bm{o}_i) \right].
\end{equation}
where $\pi_{\theta, j}(\bm{o}_i)$ is the output of the policy network at sample $\bm{o}_i$ in dimension $j$, $\sigma_i$ are independent Rademacher random variables taking values $+1$ or $-1$ with equal probability, $N$ is the number of i.i.d. data samples.

Then, the total empirical Rademacher complexity for the function class $\mathcal{F}$ is the sum over all dimensions:
\begin{equation}
    \hat{\mathfrak{R}}_N(\mathcal{F}) = \sum_{j=1}^k \hat{\mathfrak{R}}_N(\mathcal{F}_j).
\end{equation}

\begin{theorem}[Generalization Error for Policy Distillation]
    Assume that the loss function \( \ell \) is Lipschitz continuous with Lipschitz constant \( L_\ell \) with respect to its first argument, and bounded by C. Then, with probability at least \( 1 - \delta \), for all \( \theta \in \Theta \):
\[
\mathcal{L}(\pi_{\theta}) \leq \hat{\mathcal{L}}(\pi_{\theta}) + 2 L_\ell  \mathfrak{R}_N(\mathcal{F}) + C \sqrt{\frac{\log(1/\delta)}{2N}},
\]
where $\mathcal{L}(\pi_\theta) = \mathbb{E}[\ell(\pi_{\theta}, \pi^*)]$ is the expected of loss of visuomotor policy $\pi_{\theta}$ to expert policy $\pi_e$ and $ \hat{\mathcal{L}}(\pi_{\theta})$ is the empirical loss as defined in Eqn.~(\ref{equ:cloning_loss}). The Rademacher complexity  $\mathfrak{R}_N(\mathcal{F}) = \mathbb{E} [ \hat{\mathfrak{R}}_N(\mathcal{F}) ]$.
\end{theorem}

Proof is detailed in ~\citep{kakade2008complexity}. Obviously, the generalization error largely depends on the sample size. Although simply increasing $N$ does not linearly reduce error, data augmentation—specifically \textit{strong augmentation}—can effectively diversify data distribution and reduce Rademacher complexity. However, indiscriminate data augmentation across whole observations could distort crucial control information, leading to unpredictable increases in control prediction errors $ \hat{\mathcal{L}}(\pi_{\theta})$. Optimally, augmenting only control-irrelevant data while preserving essential control information can effectively reduce generalization errors by increasing the number of data samples while maintaining low prediction errors. Furthermore, direct guidance from expert policies simplifies training visuomotor policies by constraining the output space and reducing Rademacher complexity.

\begin{table*}
    \caption{\textbf{DMC-GB Generalization Performance.} 
    We report the episode returns over five random seeds. $\Delta$ represents the improvement of \ourmethod~over the second-best results. \textit{Italicized numbers} indicate average results reported using one officially pre-released model. The best performance is highlighted in bold and the second-best is underlined.}
    
    \label{tab:dmc_main_results}
    \centering
    \resizebox{\textwidth}{!}{
        \begin{NiceTabular}{c|r|ccccccccc|cc}
        \toprule
        {\bf Setting} & {\bf Easy Task}  & {\bf DrQ} & {\bf DrQ-v2} & {\bf RAD} & {\bf SODA} & {\bf SVEA} & {\bf TLDA}& {\bf VAI} & {\bf SAM+E} &{\bf SGQN} &  {\bf \ourmethod} & {\bf $\Delta$} \\
        
        \midrule
        
        &Ball in cup  &$380 \pm 188$  & $401 \pm 67$& $363 \pm 158$ & $\underline{939\pm 10}$ & $928 \pm 43$  &$ 892\pm 68$   & $909 \pm 44$& $\textit{925}$ &$889 \pm 87$  & $ \textbf{973} \pm 9$ & $+34$ \\
        
        &Cartpole  & $459 \pm 81$ &$267 \pm 26$ & $473 \pm 54$  & $742 \pm 73 $ & $772 \pm 46$  & $671\pm 57$  & $729 \pm 19$& $\textit{\underline{777}}$ &$ 770\pm 56$  & $\textbf{861}\pm6$  & $+84$\\
        
        Video &Walker walk  & $747 \pm 21$ & $196 \pm 52 $ &$608 \pm 92 $ & $771 \pm 66$  & $839 \pm 29$ & $873\pm34$ & $871 \pm 42$& $\textit{823}$ & $\underline{881} \pm 18$& $\textbf{929} \pm 17 $& $+48$\\
        
        easy&Walker stand  & $926 \pm 30$ & $487 \pm 83$& $879 \pm 64$ & $905 \pm 7$  & $947\pm 11$ &$\underline{973 \pm 6} $ & $948 \pm 12$ & $\textit{910}$ & $950 \pm 10$ & $\textbf{978} \pm 12$ & $+5$\\
        
        &Finger spin & $599 \pm 62$ &$491 \pm 35$ & $516 \pm 113$ & $783 \pm 51$ & $737 \pm 93$  & $744\pm18$ & $932 \pm 2$& $\textit{859}$ & $\textbf{947} \pm 16$ & $\underline{934} \pm 7$ & $-13$ \\

        &Cheetah run & $154 \pm 22$ &$79 \pm 8$ & $104\pm5$ & $190 \pm 48 $ & $275 \pm 3$  & $336\pm 57$ & $322 \pm 35$& $\textit{\underline{444}}$ & $ 257\pm49 $ & $\textbf{463}\pm 13 $ & $+19$\\
        
        &Finger turn & $230 \pm 29$ &$358\pm68$ & $172 \pm 63$ & $150 \pm 50$ & $197 \pm 98$  & $208\pm 35$ & $445 \pm 36$& $\textit{\underline{645}}$ & $547 \pm 267 $ & $\textbf{721}\pm 13 $ & $+76$\\
        \midrule
        &Average & $499$ &$326$ & $445$ & $640$ & $671$  & $671$ & $703$& $769$ & $749$ & $\textbf{837}$ & $+36(4.3\%)$ \\
        \midrule

        &Ball in cup & $100 \pm 40$ & $83 \pm 20$ & $98 \pm 40$ & $381 \pm 163$ & $492 \pm 100 $  & $257\pm57$  & $\textit{524}$ & $\textit{725}$ & $ \underline{857} \pm 29$  & $\textbf{944} \pm 7$ & $+87$\\
        
         & Cartpole & $136 \pm 29$ & $137 \pm 20$ & $152 \pm 29$ & $452 \pm 45$ & $401 \pm 38$ & $286\pm47$ & $\textit{378}$ & $\textit{337}$  &$\underline{537} \pm 33$  & $\textbf{623} \pm 22$ &$+86$ \\
        
        Video &Walker walk &$ 121 \pm 52 $& $87 \pm 5$& $80 \pm 10$ & $312 \pm 32$ & $521 \pm 68$  & $271\pm55$  & $\textit{\underline{823}}$ &$\textit{357}$ &$719 \pm 33$ & $\textbf{883} \pm 9$& $+60$ \\
        
        hard&Walker stand &$ 252 \pm 57 $&$234 \pm 50$& $229 \pm 45$ & $736 \pm 132$ & $840 \pm 16$ & $602\pm51$ & $\textit{\underline{931}} $& $\textit{612}$  &$842 \pm 36$ & $\textbf{941} \pm 11$ &$ +10 $\\
        
        &Finger spin & $38 \pm 13$  & $31 \pm 8$ & $39 \pm 20 $ & $309 \pm 49$& $361 \pm 25$& $241\pm29$ &  $\textit{752} $ & $\textit{\underline{772}}$ &$767 \pm 11$ &  $\textbf{798} \pm 5$ & $+26$\\
    
        &Cheetah run & $ 74 \pm 24$ &$34\pm6$ & $98 \pm 19$ & $155 \pm 16$ & $157 \pm 14$  & $90\pm27$ & $\textit{\underline{303}}$& $\textit{264}$ &$238 \pm 37$ & $\textbf{435} \pm 2$ & $+132$\\

        &Finger turn & $236 \pm 47$ &$159\pm8$ & $155 \pm 9$ & $155 \pm 16$ & $237 \pm 15$  & $104\pm18$ & $\textit{362}$& $\textit{\underline{563}}$ & $461 \pm 64$ & $\textbf{565} \pm 10$  & $+2$\\

        \midrule
        &Average & $137$ &$109$ & $122$ & $357$ & $429$  & $264$ & $581$ & $519$ & $\underline{631}$ & $\textbf{741}$ & $+110(17.4\%)$ \\
        \bottomrule

        \end{NiceTabular}
    }
\end{table*}

\section{Experiments}
\begin{figure}[t]
    \centering
    \includegraphics[width=1.0\columnwidth]{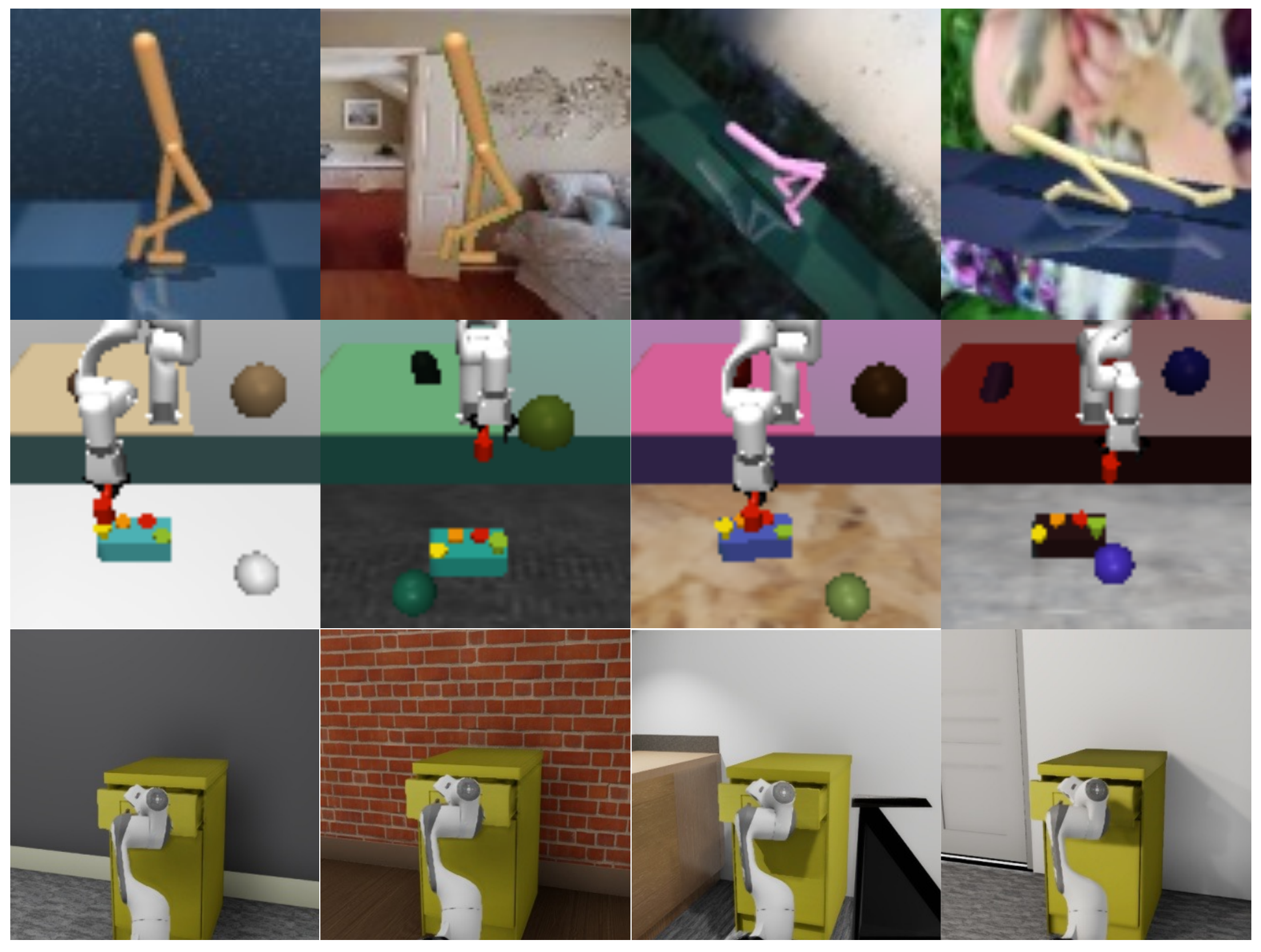}
    \caption{Observation examples from three benchmarks. Top row: DMC-GB (first: training, second: video hard setting,  last two: distraction setting). Medium row: Enhanced Robot Manipulation Distraction Benchmark (first: training, last three: testing). Bottom row: Self-designed Drawer Opening Generalization Benchmark (first: training, second: different backgrounds, last two: different scenarios). }
    \label{fig_all_env}
    \vspace{-10pt}
\end{figure}

To demonstrate the zero-shot generalization capabilities of \ourmethod, we conducted experiments across three domains, as shown in Fig.~\ref{fig_all_env}:

Firstly, we evaluated \ourmethod~ using the DMControl Generalization Benchmark (DMC-GB)~\citep{hansen2021generalization}, which tests an agent's generalization ability from a simple to a complex environment. We employed three settings here: video-easy, video-hard, and distraction. In video-hard, the background features real-world videos, significantly different from the training environment. In distraction setting, the background image, foreground color, and camera view are varied in multiple difficulty level. Each method undergoes 500k training iterations, with evaluations based only on visual inputs.

Secondly, we introduced an enhanced Robot Manipulation Distraction Benchmark (RMDB), building on the original benchmark \citep{jangir2022look,bertoin2022look} that includes four tasks: Push, Reach, Hammer, and Pegbox. We incorporated distraction objects in training, and varied colors/textures in five test environments. RMDB presents a greater challenge as it requires robots to develop a visuomotor policy robust to visual changes in all task-irrelevant areas (as shown in Fig.~\ref{fig_all_env}, where the pegbox changes to purple and brown). More detailed descriptions of RMDB can be found in Appendix~\ref{sec:supp_c}.

Finally, we design a Drawer Opening Generalization Benchmark (DOGB) that utilizes the high-fidelity NVIDIA Isaac Sim for tasks involving the opening of a drawer against large-scale background and scenario variance. A task is considered successful if the drawer opens beyond 0.29 meters within 500 time steps. Compared to RMDB, DOGB demands more precise control over the 7-DoF robot arm and 2-DoF grasp, and copes with extensive background variations (up to 400 testing environments with the change of colors, patterns, and materials of the wallpaper and the carpet).

We compared \ourmethod~to the several SOTA algorithms~\citep{yarats2021image, yarats2021mastering, laskin2020reinforcement, hansen2021generalization, hansen2021stabilizing, yuan2022don,wang2021unsupervised,bertoin2022look} in terms of generalization ability. Besides, we developed a strong baseline SAM+E that combined the large vision model SAM~\citep{kirillov2023segment} with our privilege expert. A detailed description of our settings for SAM+E and its segmentation results on DMC-GB can be found in the Appendix~\ref{sec:sam_e}. 

\subsection{Comparisons}\label{sec:compare}
\textbf{Evaluation on DMC-GB.} We conducted experiments on 7 tasks on DMC-GB following SOTA methods~\citep{bertoin2022look, wang2021unsupervised} including Ball in cup, Cartpole swingup (denoted as Cartpole), Walker walk, Walker stand, Finger spin, Cheetah run and Finger turn easy (Denoted as Finger turn). Overall, \ourmethod~achieves a significant generalization performance improvement on video-hard settings, which is $17.4\%$ higher than previous SOTA method SGQN.

We also notice that with advanced vision capabilities, SAM+E demonstrates notable generalization ability on certain tasks in the video-hard category, such as Finger Spin and Finger Turn. However, for other tasks like Walker Walk, performance in the video-hard setting declines by $56.6\%$ compared to the video-easy setting. This decrease can be attributed to the Walker environment, where the moving robot may be misclassified as background due to predefined point prompts, leading to the erroneous deletion of task-irrelevant pixels for \textit{strong augmentation}. This observation supports our intuition about applying data augmentation specifically to pixels that are irrelevant to the task. In contrast to SAM+E, our method, which incorporates a self-supervised, control-aware augmentation mask, consistently achieves high performance in both video-hard and video-easy settings. The convergence line of generalization performance on video-hard during training is shown in Fig.~\ref{fig_dmcgb-main} in the Appendix. Clearly, our method achieves higher generalization performance compared to other methods within the same number of update iterations for the visuomotor policy

\begin{figure*}[h]
    \centering
    \includegraphics[width=1.0\textwidth]{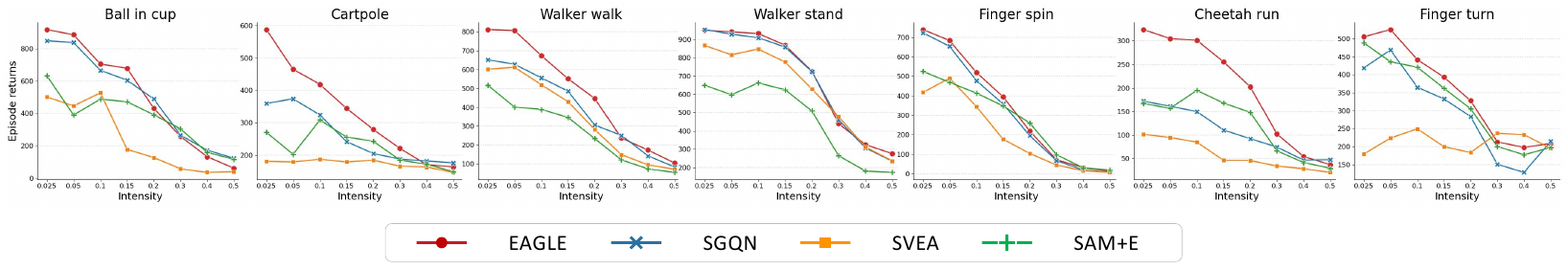}
    \caption{Generalization performance of all methods on DMC-GB distraction setting.}
    \label{fig_dmcgb-distrac}
\end{figure*}

In addition, we evaluate \ourmethod~in more challenging distracting settings~\citep{stone2021distracting} with variations in foreground color, background video, and camera view in DMC-GB. As shown in Fig.~\ref{fig_dmcgb-distrac}, \ourmethod~consistently outperforms other strong baselines across all tasks. Particularly, in Cheetah Run, \ourmethod~achieves an average return of $301$ under an intensity of $0.1$, which is $68.9\%$ higher than SAM+E (with a score of 195) and more than twice that of SGQN (with a score of 149). This improvement is attributable to the control-aware augmentation mask used by \ourmethod, which augments all task-irrelevant regions, including some parts of the foreground, to enhance generalization ability. We also note that although SAM+E achieves high performance in some tasks on video-hard, it has difficulty generalizing against the foreground and camera views with purely visual segmentation.

\textbf{Evaluation on RMDB.}
Table~\ref{tab:robo_main_results} shows the training and evaluation results of \ourmethod~and four other strong baselines on RMDB. Clearly, \ourmethod~achieves high performance in both the training and testing environments. For example, in the PegBox training task, \ourmethod~realizes a significant improvement—almost three times—in success rate compared to SGQN. Although SAM+E also achieves high performance in training, its generalization ability can be limited in some task. Specifically, it achieves $37\%$ success rate in the testing environment for PegBox, which is $26\%$ lower than that obtained by \ourmethod. With the reliable guidance of privileged Expert and a clear visual mask, SAM+E more easily learns a stable visuomotor control policy compared to other VRL methods. However, the visual mask is insufficient to handle visual variations from distracting objects, which are common in many robotic tasks. In contrast, the control-aware mask used by \ourmethod~augments the task-irrelevant regions in the observation space while preserving control-related information for decision-making.

\begin{figure}[t]
    \centering
    \includegraphics[width=1.0\columnwidth]{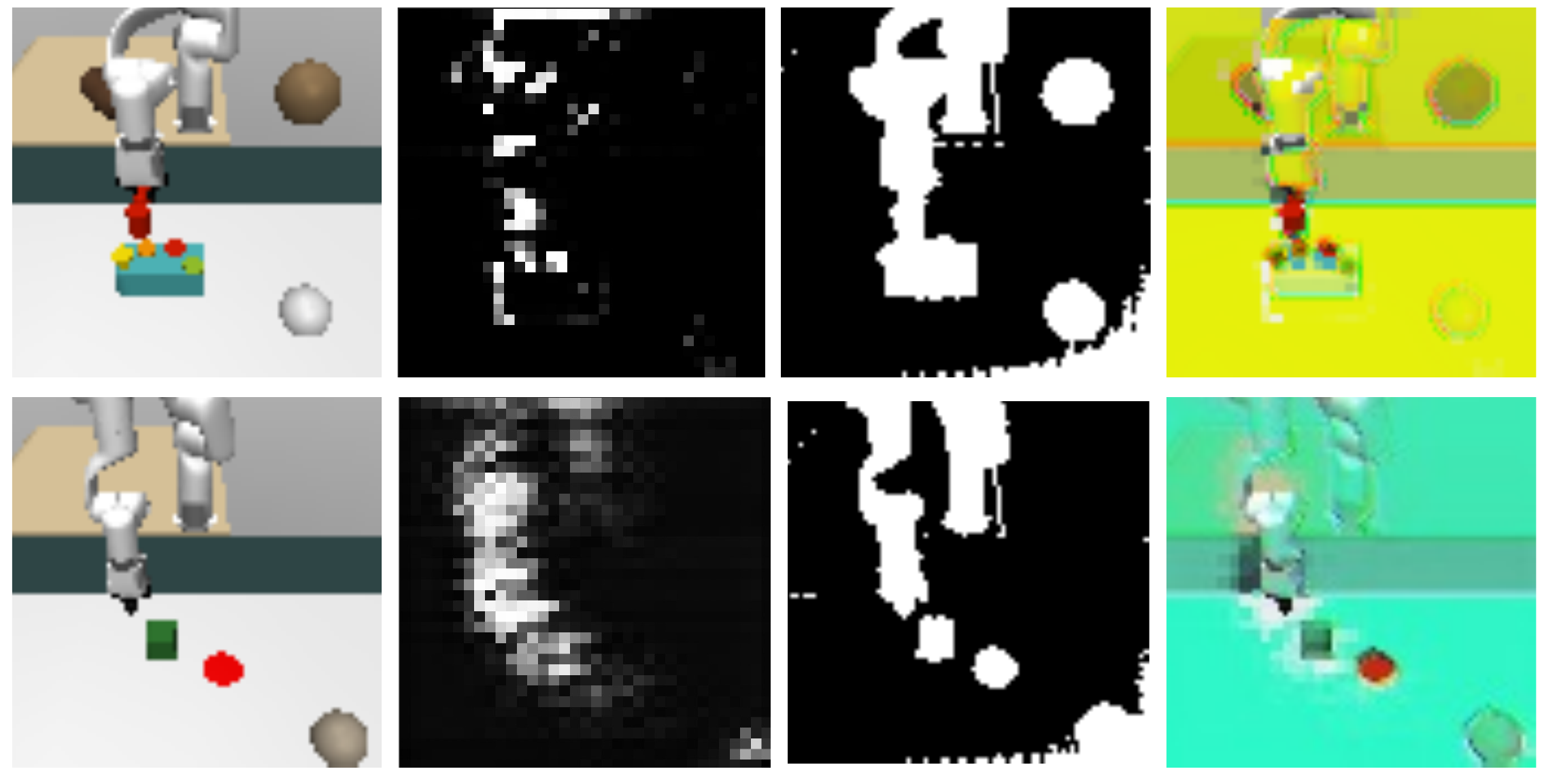}
    \caption{Illustration of the mask obtained by \ourmethod~ and SAM on Hammer and Push. First column: original observations. Second column: \ourmethod's mask. Third column: SAM's mask. Last column: control-aware augmented observations.}
    \label{fig:robot-demo}
\end{figure}

In Fig.~\ref{fig:robot-demo}, we visualize the masks obtained by \ourmethod~and SAM for the tasks Hammer and Push. For Hammer, SAM identifies the entire robot arm, the hammer, the toy box, and two distracting balls as foreground, treating other regions as background. In contrast, \ourmethod~only captures parts of essential joints/links in the robot arm, the hammer, and the button in the toy box with the mask, augmenting other regions with \textit{random conv}, as shown in the second and fourth pictures in the top row of Fig.~\ref{fig:robot-demo}. This selective preservation is enabled by the auxiliary losses in Eqn.~(\ref{equ:att_loss}), which ensures that only the most control-related areas are maintained for control prediction. It is important to note that \ourmethod~can capture all control-related regions, regardless of whether they are dynamic or static. For example, in Push task, the robot must push a green box towards a static red target throughout an episode, yet \ourmethod~still captures this for minimizing control prediction loss. Thus, with essential goal information carefully preserved, the training and testing performance of \ourmethod~are both superior compared to other baselines. More visualization results on RMDB and our detailed settings of SAM can be found in the appendix.

\begin{table}
    \caption{Success rate on Robot Manipulation Generalization Performance. Test success rate denotes the average results across 5 different test environments. 
    }
    \label{tab:robo_main_results}
    \centering
    \resizebox{0.9\columnwidth}{!}{
        \begin{NiceTabular}{c|cc|cc}
        \toprule
        {\bf Method} & \multicolumn{2}{c}{\bf Hammer}  & \multicolumn{2}{c}{\bf Pegbox}  \\
        \midrule
          &Train & Test & Train & Test \\
        \midrule

        {\bf SVEA} & $ 0.02$  & $ 0.04 \pm 0.02 $  & $ 0.42$ & $0.03 \pm 0.02$\\
        {\bf TLDA} & $0.22$  & $0.02\pm0.03$  & $0.00$ & $0.00\pm 0.01$\\
        {\bf VAI} & $0.00$  & $0.01\pm0.01$  & $0.00$ & $0.00\pm 0.00$\\
        {\bf SAM+E} & $0.46$  & $0.06\pm0.05$  & $0.50$ & $0.50\pm 0.15$\\
        {\bf SGQN} & $0.49$  & $0.10\pm0.12$  & $0.21$ & $0.05\pm 0.03$\\
        \midrule
        {\bf \ourmethod} & $\textbf{0.50}$  & $\textbf{0.21}\pm0.06$  & $\textbf{0.58}$ & $\textbf{0.54}\pm 0.20$\\
        \midrule
        & \multicolumn{2}{c}{\bf Push} & \multicolumn{2}{c}{\bf{Reach}}  \\
        \midrule
          &Train & Test & Train & Test \\
        \midrule

        {\bf SVEA} & $0.30$ & $ 0.30 \pm 0.00 $ & $0.98$ & $0.78 \pm 0.23$\\
        {\bf TLDA} & $0.34$ & $0.32\pm0.04$ & $0.98$ & $0.96\pm 0.03$\\
        {\bf VAI} & $0.23$ & $ 0.21\pm 0.02$ & $0.80$ & $0.36\pm 0.19$\\
        {\bf SAM+E} & $0.68$ & $0.37\pm0.07$ & $1.00$ & $0.99\pm 0.01$\\
        {\bf SGQN} & $0.38$ & $0.37\pm 0.01$ & $1.00$ & $0.98 \pm 0.03$\\
        \midrule
        {\bf \ourmethod} & $\textbf{0.72}$ & $ \textbf{0.50}\pm 0.10 $ & $\textbf{1.00}$ & $\textbf{1.00}\pm 0.00$\\
                
        \bottomrule
        \end{NiceTabular}
    }
\end{table}

\subsection{Ablation studies}

\begin{table*}
    \caption{\textbf{Ablation study of our control-aware attention module (denoted as Att.) , privilege-guided distillation module (denoted as Exp.) and the \textit{random overlay} augmentation (denoted as Aug.) on DMC-GB.}  
    We designed four ablation methods for comparison. 1) \textbf{Q-only}, 2) \textbf{Q+Aug}, 3) \textbf{Q+Mask} and 4) \textbf{E+Aug}.
    }
    \label{tab:dmc_ablation}
    \centering
    \resizebox{1.0\textwidth}{!}{
        \begin{NiceTabular}{c|l|cccc|cccccc|c}
        \toprule
       Setting& & Q.& Aug. & Att. & Exp.  & {\bf Ball in cup} & {\bf Walker walk} & {\bf Walker stand}& {\bf Finger spin}  & {\bf Finger turn} & {\bf Cheetah run} & {\bf Average}\\
        \midrule
        
        &\textbf{Q-only} & \ding{51} & & & & $678.1 \pm 387$ & $467.7 \pm120$ & $872.7 \pm 92.1$ & $839.3 \pm 51$  & $454.2\pm 300$ & $457.3 \pm 293.8$& $628.3$ \\

        &\textbf{Q+Aug} & \ding{51} & \ding{51}& & & $766.8\pm203(
        +\textbf{13}\%)$ & $435.6 \pm 29(-\textbf{7}\%)$ & $473.2 \pm 331(-\textbf{46}\%)$ & $658.4 \pm 29(-\textbf{22}\%)$& $246.8 \pm 41(-\textbf{45.7}\%) $ & $227.6\pm 92(-\textbf{50.2}\%)$ & $468.0$\\
        
        Train &\textbf{Q+Mask}& \ding{51} & \ding{51} & \ding{51} &   & $801.1 \pm 120(\textbf{+18}\%)$ & $673.1 \pm 185(+\textbf{44}\%) $ & $965.6 \pm 7(+\textbf{11}\%)$ & $844.6 \pm 22(\textbf{0}\%)$ &$679.2\pm 16(\textbf{49}\%)$& $256.8 \pm 12(-\textbf{44}\%)$& $702.9$\\

       &\textbf{E+Aug} &  & \ding{51} &  & \ding{51} & $970.2 \pm 2 (+\textbf{43}\%)$ &  $931.4 \pm 34(+\textbf{99}\%)$  &$907.4 \pm 25(+\textbf{4}\%)$ & $843.5 \pm 23 (+\textbf{1}\%)$ & $914.6 \pm 55(+\textbf{101.4}\%)$ & $393.9 \pm 11(-\textbf{13.9}\%)$ & $826.9$\\
       
        &\textbf{\ourmethod} & & \ding{51} & \ding{51} & \ding{51} &$968.2 \pm 8(+\textbf{43}\%)$   & $946.4 \pm 5(+\textbf{102}\%)$ &$952.1 \pm 15(+\textbf{9}\%) $ & $839.3 \pm 8(+\textbf{0}\%)$  &$788.0 \pm 47(+\textbf{73}\%)$ & $838.8 \pm 15(+\textbf{83}\%)$ &$888.8$ \\

        \midrule
        
        &\textbf{Q-only} & \ding{51} & & & &  $401.3 \pm 67$ & $195.7 \pm 52$  &  $487.0 \pm 83$  & $490.6 \pm 35$ & $258.3 \pm 68$ & $78.5\pm 8$ &$318.6$\\
        
        &\textbf{Q+Aug} & \ding{51} & \ding{51}& & & $722.9 \pm 169.3(+\textbf{80}\%)$ & $369.7 \pm 12(+\textbf{89}\%)$ & $478.6 \pm 329.3(-\textbf{2}\%)$ & $586.5 \pm 20(+\textbf{20}\%)$  & $212.9 \pm 85(-\textbf{18}\%)$  & $212.6 \pm 62(+\textbf{171}\%) $ &$430.60$\\
        
        Video &\textbf{Q+Mask}& \ding{51} & \ding{51} & \ding{51} &   & $724.7 \pm 107(+\textbf{81}\%)$ & $633.2 \pm 174(+\textbf{239}\%)$ & $961.3 \pm 2(+\textbf{239}\%)$ & $767.9 \pm 19(+\textbf{239}\%)$    &$298.2\pm18(\textbf{16}\%)$ & $296.4 \pm 53(+\textbf{278}\%)$& $613.0$\\

       easy&\textbf{E+Aug} &  & \ding{51} &  & \ding{51} & $ 954.8 \pm 12(+\textbf{138}\%)$ & $828.1 \pm 59(+\textbf{323}\%)$  &$ 886.2 \pm 41(+\textbf{82}\%)$ &$741.0 \pm 8(+\textbf{51}\%)$ & $577.8 \pm 13(+\textbf{124}\%)$ & $325.4 \pm 11(+\textbf{314}\%)$ & $718.9$\\
       
        &\textbf{\ourmethod} & & \ding{51} & \ding{51} & \ding{51} &$973.4 \pm 9(+\textbf{143}\%)$ & $929.3 \pm 17(+\textbf{375}\%)$ &$ 978.5 \pm 12(+\textbf{101}\%)$ & $934.1 \pm 7(+\textbf{70}\%)$  &$721.3\pm13(+\textbf{179}\%)$ & $463.2\pm13(+\textbf{490}\%)$ & $833.3$ \\

        \midrule
        
        &\textbf{Q-only} & \ding{51} & & & &  $83.4 \pm 20$   &$87.4 \pm 5$   &  $233.5 \pm 50$ &  $30.7 \pm 8$ &$158.9 \pm 8$ & $34.1 \pm 6$ & $104.7$\\

        &\textbf{Q+Aug} & \ding{51} & \ding{51}& & & $313.1 \pm 108(+\textbf{271}\%)$ & $233.3 \pm 34(+\textbf{167}\%)$ & $382.9 \pm 245(+\textbf{64}\%)$ & $276.6 \pm 15(+\textbf{801}\%)$  & $225.2 \pm 62(+\textbf{42}\%) $ & $113.0 \pm 10(+\textbf{231}\%)$& $257.3$\\
        
        Video &\textbf{Q+Mask}& \ding{51} & \ding{51} & \ding{51} &   & $459.5 \pm 31(+\textbf{451}\%)$ & $483.4 \pm 136(+\textbf{453}\%)$ & $910.4 \pm 17(+\textbf{290}\%)$& $724.9 \pm 10(+\textbf{2216}\%)$    &$294.2\pm31(+\textbf{85}\%)$& $183.2 \pm 11(+\textbf{437}\%)$& $509.3$\\

       hard&\textbf{E+Aug} &  & \ding{51} &  & \ding{51} &  $657.4\pm28(+\textbf{688}\%)$  & $457.8 \pm 44(+\textbf{424}\%)$ &  $666.4 \pm 13(+\textbf{185}\%)$  &$363.9\pm9(+\textbf{1085}\%)$ & $394.5\pm 48(+\textbf{149}\%)$ & $104.9 \pm 12(+\textbf{208}\%)$ & $440.5$\\
       
        &\textbf{\ourmethod} & & \ding{51} & \ding{51} & \ding{51} &$ 944.0 \pm 7(+\textbf{1032}\%)$  & $882.5\pm 75(+\textbf{908}\%)$ &$941.3 \pm 11(+\textbf{303}\%)$ & $797.9 \pm 5(+\textbf{2499}\%)$ & $565.3\pm10(+\textbf{256}\%)$ &$435.1 \pm 2(+\textbf{1179}\%)$  &$761.3$ \\
        \bottomrule
        \end{NiceTabular}
    }
\end{table*}

We further study how each component of \ourmethod~contributes to its generalization and stability. In this subsection, we first conduct ablation studies of the control-aware augmentation module and the privilege-guided distillation module on DMC-GB and DOGB. Then, we investigate the extent of improvement that the privilege Expert brings to other state-of-the-art (SOTA) methods on challenging RMDB tasks for these VRL methods. Lastly, we perform ablations on the loss function in Eqn.~(\ref{equ:att_loss}) on DMC-GB.

\textbf{Ablation on Module Structure:} We conducted extensive ablation studies on \ourmethod~to evaluate the effectiveness of the control-aware data augmentation module and the privilege-guided distillation module. As shown in Tab.~\ref{tab:dmc_ablation}, we designed four ablation conditions for comparison:
1) \textbf{Q-only}: the vanilla DrQ-v2 algorithm with random shift augmentation.
2) \textbf{Q+Aug}: this method applies \textit{random overlay} augmentation to \textbf{Q-only}.
3) \textbf{Q+Mask}: this method adds a control-aware attention mask to \textbf{Q+Aug}. It can be considered as \ourmethod~w/o. Expert module.
4) \textbf{E+Aug}: this method utilizes privilege-guided policy distillation and applies \textit{random overlay} to the input image. It can be considered as \ourmethod~w/o. Mask module.

In Tab.~\ref{tab:dmc_ablation}, we find that applying strong augmentation directly to the DRL baseline (Q-only) can degrade policy performance, even in the training environment. For example, in the Walker walk task, Q+Aug achieved a training score of $435.6$, which is $7\%$ lower than Q-only. In contrast, as demonstrated by Q+Mask and E+Aug, both the control-aware attention mask and the privilege expert utilize overlay augmentation to enhance training and generalization performance on DMC-GB. Specifically, Q+Mask and E+Aug achieved average increases of $44\%$ and $99\%$ in training performance, respectively, and $239\%$ and $323\%$ in the video-easy environment.

It is worth noting that in most tasks under the video-hard setting, the control-aware augmentation module significantly improves generalization performance, whereas the privilege expert module reduces performance variability. For example, in Walker Walk, Q+Mask resulted in a $453\%$ increase over Q-only, but with a high variance of 136. In contrast, E+Aug improved generalization with a much lower variance of $44$. Therefore, by integrating the control-aware attention mask and privilege-guided distillation module, \ourmethod~achieves a remarkable improvement of $908\%$ in Walker Walk. This underscores the joint effect of two modules in enhancing the efficient generalization of visuomotor policies.

\begin{figure}
\centering
    \begin{subfigure}{0.45\columnwidth}
            \includegraphics[height=3cm]{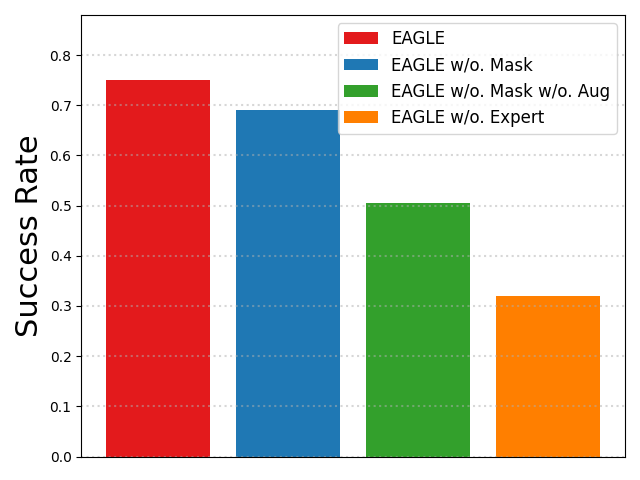}
            \caption{Train environment}
    \end{subfigure}%
    \hspace{2pt}
    \begin{subfigure}{0.45\columnwidth}
            \includegraphics[height=3cm]{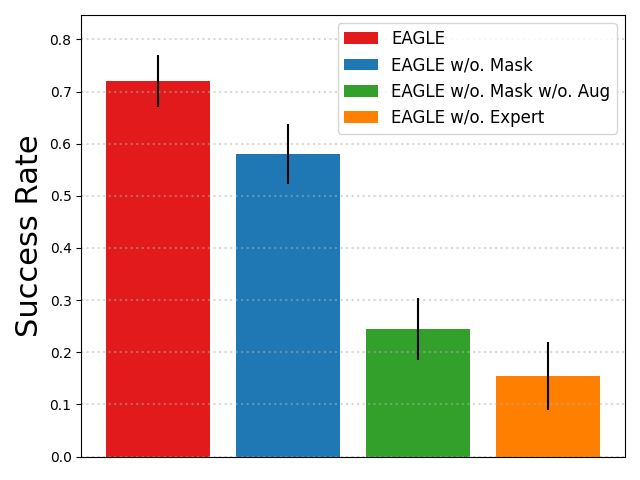}
            \caption{Test environment}
    \end{subfigure}
    \caption{Ablation performance on DOGB.}
    \label{fig:dogb-result}
\end{figure}

\begin{figure}[t]
    \centering
    \includegraphics[width=1.0\columnwidth]{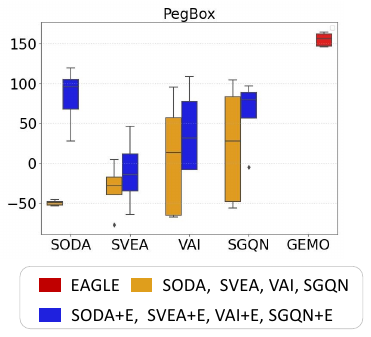}
    \caption{Improvements on baselines with privilege expert.}
    \label{fig:exper-ablation}
\end{figure}

We conducted ablation studies to validate our method on a long-horizontal drawer opening task in DOGB. As shown in Fig.~\ref{fig:dogb-result}, when facing unseen backgrounds, \ourmethod~consistently achieved an average success rate of $72\%$ across large-scale testing environments. In contrast, \ourmethod~w/o. Expert showed significantly lower performance, with success rates of $32\%$ and $16\%$ in the training and testing environments, respectively, demonstrating the importance of the privileged Expert in handling more challenging tasks. Additionally, while \ourmethod~w/o. Mask performed comparably to the \ourmethod~in the training environment, it exhibited poor generalization in the testing environment, highlighting the contribution of our attention mask for generalization.

\textbf{Combine Expert module with other baselines.} We investigated whether the Expert module could consistently enhance the generalization performance across other baselines besides SAM. We integrated the Expert with SODA, SVEA, VAI, and SGQN on RMDB, considering the challenging Pegbox task for VRL methods on this benchmark. As shown in Fig.~\ref{fig:exper-ablation}, the Expert module notably improved the generalization ability of other baselines. For example, SGQN+E achieved almost a two-times increase in the average returns, from $22$ to $63$, after incorporating the Expert module, along with a $44.6\%$ reduction in variance. This demonstrates the importance of leveraging privileged information in the training environment when addressing challenging robotic tasks. Additionally, we observed that after integration with the Expert, \ourmethod~continued to outperform other methods, attributable to the control-aware augmentation module in \ourmethod~that selectively applies augmentation to task-irrelevant regions, thus consistently enhancing generalization performance during training.

\begin{figure}[t]
    \centering
    \includegraphics[width=1.0\columnwidth]{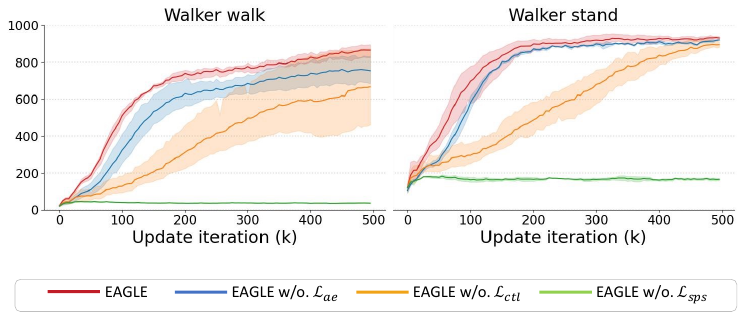}
    \caption{Ablation study of our control prediction loss and the auto-encoder loss in our augmentation module.}
    \label{fig:ablation-loss}
\end{figure}

\textbf{Ablation on the loss function.} In Eqn.~\ref{equ:att_loss}, we formulate a self-supervised reconstruction task along with three auxiliary losses to derive a control-aware mask for augmentation. We conducted ablation studies of these three losses on DMC-GB. The convergence results for the video-hard settings of Walker walk and Walker stand are shown in Fig.~\ref{fig:ablation-loss}, with additional results available in the appendix. Notably, the sparsity loss is crucial for generalization; otherwise, the generalization policy may collapse. Then, we can observe that the control prediction loss, $\mathcal{L}_{ctl}$, significantly enhances sample efficiency. For example, in Walker Stand, \ourmethod~rapidly converged to a generalizable visuomotor policy within 200k training iterations, whereas \ourmethod~without $\mathcal{L}_{ctl}$ only reached similar performance levels at 500k update iterations and exhibited higher variance. This efficiency stems from $\mathcal{L}_{ctl}$'s ability to focus the model on the most critical regions for decision-making, given the constraints of the limited mask region. Additionally, the auto-encoder loss $l_{ae}$ impacts tasks differently. In Walker walk, overlooking $l_{ae}$ leads to high variance and unstable generalization performance. Conversely, in Walker stand, its influence on stability is minimal.

\subsection{Efficiency comparison}
\textbf{Mask generating time efficiency}. We evaluated the efficiency of \ourmethod~ relative to four other methods by measuring the time required for mask generation in augmentation or consistency learning. Specifically, VAI~\citep{wang2021unsupervised} uses a reconstruction structure similar to \ourmethod~, but it generates masks through a decoder network using a predefined threshold. SGQN~\citep{bertoin2022look} produces sharp saliency maps by calculating policy gradients, which are then used in a consistency loss. TLDA~\cite{yuan2022don} derives masks by computing the Lipschitz constant based on policy changes before and after image perturbation. SAM~\citep{kirillov2023segment} generates masks based on given point prompts. Although these methods only compute masks during training, the inference time for masks can significantly affect training efficiency due to the large observation space. We computed the average mask inference time (in milliseconds) per image over 15,000 iterations. All evaluations were conducted on an Ubuntu 18.04 server equipped with an 8GB NVIDIA GTX 1180 graphics card and a 12-core AuthenticAMD CPU. As shown in Tab.~\ref{tab:time_efficiency}, \ourmethod~demonstrates time efficiency in mask generation, contributing to the lightweight encoder in our implementation and the direct usage of the latent attention mask without decoding.

\begin{table}
    \caption{\textbf{Efficiency comparison of \ourmethod~vs. other methods regarding mask generation.}  
    }
    \label{tab:time_efficiency}
    \centering
    \resizebox{0.8\columnwidth}{!}{
        \begin{NiceTabular}{c|ccccc}
        \toprule
        {\bf Method} & {\bf \ourmethod}  & {\bf VAI}& {\bf SGQN}& {\bf TLDA} & {\bf SAM}  \\
        \midrule

        {\bf time (ms) per image} $\downarrow$ & $1.028$ & $4.989$& $8.692$  & $10.668$ & $329.941$  \\
                  
        \bottomrule
        \end{NiceTabular}
    }
\end{table}

\textbf{Training efficiency}. We assessed the training efficiency of \ourmethod~in comparison with other SOTA methods, as detailed in Tab.~\ref{tab:train_efficiency_dmcgb}. The average time efficiency across seven tasks in DMC-GB is provided. \textbf{Expert} denotes the privileged expert utilized in \ourmethod. Notably, \ourmethod~demonstrates high training efficiency, characterized by both reduced training times and enhanced generalization performance. For example, while SGQN achieved well generalization performance in DMC-GB, it requires twice the training time of \ourmethod. Despite requiring similar training times as SVEA and SODA, \ourmethod~achieves a remarkable average reward of 741 in the DMC-GB video-hard settings, which is $72.4\%$ and $107.6\%$ higher than that achieved by SVEA and SODA, respectively, as shown in Tab.~\ref{tab:dmc_main_results}.

\begin{table}
    \caption{\textbf{Training Efficiency comparison of \ourmethod~vs. other SOTAs.}  
    }
    \label{tab:train_efficiency_dmcgb}
    \centering
    \resizebox{0.9\columnwidth}{!}{
        \begin{NiceTabular}{c|ccccccc}
        \toprule
        {\bf Method} & {\bf Expert} & {\bf \ourmethod}  &  {\bf SVEA}& {\bf SODA} &{\bf SGQN} &  {\bf SAM+E} &  {\bf TLDA}\\
        \midrule

        {\bf training time (h)} $\downarrow$ & $0.9$ & $8.3$& $9.0$ & $8.0$ & $16.3$& $22.8$ & $61.3$  \\
                  
        \bottomrule
        \end{NiceTabular}
    }
\end{table}

\section{Conclusion}
In this paper, we address the generalization challenge of visuomotor policies in the face of visual changes. We propose an \textbf{E}fficient tr\textbf{A}ining framework for \textbf{G}enera\textbf{L}izabl\textbf{E} visuomotor policies (\ourmethod) designed to identify control-related regions and facilitate zero-shot generalization to unseen environments. \ourmethod~comprises two jointly optimized modules: a control-aware augmentation module and a privilege-guided policy distillation module. The former leverages a self-supervised reconstruction task with three auxiliary losses to learn a control-aware attention mask, which distinguishes task-irrelevant pixels and applies \textit{strong augmentations} to minimize generalization gaps. The latter distills knowledge from a pretrained privileged expert into the visuomotor policies. We conduct extensive comparative and ablation studies across three challenging benchmarks to assess the efficacy of \ourmethod. The experimental results well validate the effectiveness of our approach.

{
    \small
    \bibliographystyle{ieeenat_fullname}
    \bibliography{main}

\begin{thebibliography}{60}
\providecommand{\natexlab}[1]{#1}
\providecommand{\url}[1]{\texttt{#1}}
\expandafter\ifx\csname urlstyle\endcsname\relax
  \providecommand{\doi}[1]{doi: #1}\else
  \providecommand{\doi}{doi: \begingroup \urlstyle{rm}\Url}\fi

\bibitem[Banino et~al.(2018)Banino, Barry, Uria, Blundell, Lillicrap, Mirowski,
  Pritzel, Chadwick, Degris, Modayil, et~al.]{banino2018vector}
Andrea Banino, Caswell Barry, Benigno Uria, Charles Blundell, Timothy
  Lillicrap, Piotr Mirowski, Alexander Pritzel, Martin~J Chadwick, Thomas
  Degris, Joseph Modayil, et~al.
\newblock Vector-based navigation using grid-like representations in artificial
  agents.
\newblock \emph{Nature}, 557\penalty0 (7705):\penalty0 429--433, 2018.

\bibitem[Bertoin et~al.(2022)Bertoin, Zouitine, Zouitine, and
  Rachelson]{bertoin2022look}
David Bertoin, Adil Zouitine, Mehdi Zouitine, and Emmanuel Rachelson.
\newblock Look where you look! saliency-guided q-networks for visual rl tasks.
\newblock \emph{Advances in neural information processing systems}, 2022.

\bibitem[Chen et~al.(2020{\natexlab{a}})Chen, Zhou, Koltun, and
  Kr{\"a}henb{\"u}hl]{chen2020learning}
Dian Chen, Brady Zhou, Vladlen Koltun, and Philipp Kr{\"a}henb{\"u}hl.
\newblock Learning by cheating.
\newblock In \emph{Conference on Robot Learning}, pages 66--75. PMLR,
  2020{\natexlab{a}}.

\bibitem[Chen et~al.(2020{\natexlab{b}})Chen, Ghadirzadeh, Bj{\"o}rkman, and
  Jensfelt]{chen2020adversarial}
Xi Chen, Ali Ghadirzadeh, M{\aa}rten Bj{\"o}rkman, and Patric Jensfelt.
\newblock Adversarial feature training for generalizable robotic visuomotor
  control.
\newblock In \emph{2020 IEEE International Conference on Robotics and
  Automation (ICRA)}, pages 1142--1148. IEEE, 2020{\natexlab{b}}.

\bibitem[Choi et~al.(2023)Choi, Lee, Jeong, and Min]{choi2023environment}
Hyesong Choi, Hunsang Lee, Seongwon Jeong, and Dongbo Min.
\newblock Environment agnostic representation for visual reinforcement
  learning.
\newblock In \emph{Proceedings of the IEEE/CVF International Conference on
  Computer Vision}, pages 263--273, 2023.

\bibitem[Cobbe et~al.(2019)Cobbe, Klimov, Hesse, Kim, and
  Schulman]{cobbe2019quantifying}
Karl Cobbe, Oleg Klimov, Chris Hesse, Taehoon Kim, and John Schulman.
\newblock Quantifying generalization in reinforcement learning.
\newblock In \emph{International Conference on Machine Learning}, pages
  1282--1289. PMLR, 2019.

\bibitem[Cobbe et~al.(2020)Cobbe, Hesse, Hilton, and
  Schulman]{cobbe2020leveraging}
Karl Cobbe, Chris Hesse, Jacob Hilton, and John Schulman.
\newblock Leveraging procedural generation to benchmark reinforcement learning.
\newblock In \emph{International conference on machine learning}, pages
  2048--2056. PMLR, 2020.

\bibitem[Czarnecki et~al.(2019)Czarnecki, Pascanu, Osindero, Jayakumar,
  Swirszcz, and Jaderberg]{czarnecki2019distilling}
Wojciech~M Czarnecki, Razvan Pascanu, Simon Osindero, Siddhant Jayakumar,
  Grzegorz Swirszcz, and Max Jaderberg.
\newblock Distilling policy distillation.
\newblock In \emph{The 22nd international conference on artificial intelligence
  and statistics}, pages 1331--1340. PMLR, 2019.

\bibitem[Dosovitskiy et~al.(2017)Dosovitskiy, Ros, Codevilla, Lopez, and
  Koltun]{dosovitskiy2017carla}
Alexey Dosovitskiy, German Ros, Felipe Codevilla, Antonio Lopez, and Vladlen
  Koltun.
\newblock Carla: An open urban driving simulator.
\newblock In \emph{Conference on robot learning}, pages 1--16. PMLR, 2017.

\bibitem[Fan and Li(2022)]{fan2022dribo}
Jiameng Fan and Wenchao Li.
\newblock Dribo: Robust deep reinforcement learning via multi-view information
  bottleneck.
\newblock In \emph{International Conference on Machine Learning}, pages
  6074--6102. PMLR, 2022.

\bibitem[Fan et~al.(2021)Fan, Wang, Huang, Yu, Fei-Fei, Zhu, and
  Anandkumar]{fan2021secant}
Linxi Fan, Guanzhi Wang, De-An Huang, Zhiding Yu, Li Fei-Fei, Yuke Zhu, and
  Animashree Anandkumar.
\newblock Secant: Self-expert cloning for zero-shot generalization of visual
  policies.
\newblock In \emph{International Conference on Machine Learning}, pages
  3088--3099. PMLR, 2021.

\bibitem[Fu et~al.(2021)Fu, Yang, Agrawal, and Jaakkola]{fu2021learning}
Xiang Fu, Ge Yang, Pulkit Agrawal, and Tommi Jaakkola.
\newblock Learning task informed abstractions.
\newblock In \emph{International Conference on Machine Learning}, pages
  3480--3491. PMLR, 2021.

\bibitem[Galashov et~al.(2022)Galashov, Merel, and Heess]{galashov2022data}
Alexandre Galashov, Josh~S Merel, and Nicolas Heess.
\newblock Data augmentation for efficient learning from parametric experts.
\newblock \emph{Advances in Neural Information Processing Systems},
  35:\penalty0 31484--31496, 2022.

\bibitem[Haarnoja et~al.(2018)Haarnoja, Zhou, Abbeel, and
  Levine]{haarnoja2018soft}
Tuomas Haarnoja, Aurick Zhou, Pieter Abbeel, and Sergey Levine.
\newblock Soft actor-critic: Off-policy maximum entropy deep reinforcement
  learning with a stochastic actor.
\newblock In \emph{International conference on machine learning}, pages
  1861--1870. PMLR, 2018.

\bibitem[Haldar et~al.(2023)Haldar, Mathur, Yarats, and Pinto]{haldar2023watch}
Siddhant Haldar, Vaibhav Mathur, Denis Yarats, and Lerrel Pinto.
\newblock Watch and match: Supercharging imitation with regularized optimal
  transport.
\newblock In \emph{Conference on Robot Learning}, pages 32--43. PMLR, 2023.

\bibitem[Hansen and Wang(2021)]{hansen2021generalization}
Nicklas Hansen and Xiaolong Wang.
\newblock Generalization in reinforcement learning by soft data augmentation.
\newblock In \emph{2021 IEEE International Conference on Robotics and
  Automation (ICRA)}, pages 13611--13617. IEEE, 2021.

\bibitem[Hansen et~al.(2021)Hansen, Su, and Wang]{hansen2021stabilizing}
Nicklas Hansen, Hao Su, and Xiaolong Wang.
\newblock Stabilizing deep q-learning with convnets and vision transformers
  under data augmentation.
\newblock \emph{Advances in neural information processing systems},
  34:\penalty0 3680--3693, 2021.

\bibitem[Hern{\'a}ndez-Garc{\'\i}a and K{\"o}nig(2018)]{hernandez2018data}
Alex Hern{\'a}ndez-Garc{\'\i}a and Peter K{\"o}nig.
\newblock Data augmentation instead of explicit regularization.
\newblock \emph{arXiv preprint arXiv:1806.03852}, 2018.

\bibitem[Igl et~al.(2019)Igl, Ciosek, Li, Tschiatschek, Zhang, Devlin, and
  Hofmann]{igl2019generalization}
Maximilian Igl, Kamil Ciosek, Yingzhen Li, Sebastian Tschiatschek, Cheng Zhang,
  Sam Devlin, and Katja Hofmann.
\newblock Generalization in reinforcement learning with selective noise
  injection and information bottleneck.
\newblock \emph{Advances in neural information processing systems}, 32, 2019.

\bibitem[Igl et~al.(2020)Igl, Farquhar, Luketina, Boehmer, and
  Whiteson]{igl2020impact}
Maximilian Igl, Gregory Farquhar, Jelena Luketina, Wendelin Boehmer, and Shimon
  Whiteson.
\newblock The impact of non-stationarity on generalisation in deep
  reinforcement learning.
\newblock \emph{arXiv preprint arXiv:2006.05826}, 2020.

\bibitem[Jangir et~al.(2022)Jangir, Hansen, Ghosal, Jain, and
  Wang]{jangir2022look}
Rishabh Jangir, Nicklas Hansen, Sambaran Ghosal, Mohit Jain, and Xiaolong Wang.
\newblock Look closer: Bridging egocentric and third-person views with
  transformers for robotic manipulation.
\newblock \emph{IEEE Robotics and Automation Letters}, 7\penalty0 (2):\penalty0
  3046--3053, 2022.

\bibitem[Justesen et~al.(2018)Justesen, Torrado, Bontrager, Khalifa, Togelius,
  and Risi]{justesen2018illuminating}
Niels Justesen, Ruben~Rodriguez Torrado, Philip Bontrager, Ahmed Khalifa,
  Julian Togelius, and Sebastian Risi.
\newblock Illuminating generalization in deep reinforcement learning through
  procedural level generation.
\newblock \emph{NeurIPS Workshop on Deep Reinforcement Learning Workshop},
  2018.

\bibitem[Kakade et~al.(2008)Kakade, Sridharan, and
  Tewari]{kakade2008complexity}
Sham~M Kakade, Karthik Sridharan, and Ambuj Tewari.
\newblock On the complexity of linear prediction: Risk bounds, margin bounds,
  and regularization.
\newblock \emph{Advances in neural information processing systems}, 21, 2008.

\bibitem[Kalashnikov et~al.(2018)Kalashnikov, Irpan, Pastor, Ibarz, Herzog,
  Jang, Quillen, Holly, Kalakrishnan, Vanhoucke,
  et~al.]{kalashnikov2018scalable}
Dmitry Kalashnikov, Alex Irpan, Peter Pastor, Julian Ibarz, Alexander Herzog,
  Eric Jang, Deirdre Quillen, Ethan Holly, Mrinal Kalakrishnan, Vincent
  Vanhoucke, et~al.
\newblock Scalable deep reinforcement learning for vision-based robotic
  manipulation.
\newblock In \emph{Conference on Robot Learning}, pages 651--673. PMLR, 2018.

\bibitem[Kemertas and Aumentado-Armstrong(2021)]{kemertas2021towards}
Mete Kemertas and Tristan Aumentado-Armstrong.
\newblock Towards robust bisimulation metric learning.
\newblock \emph{Advances in Neural Information Processing Systems},
  34:\penalty0 4764--4777, 2021.

\bibitem[Kirillov et~al.(2023)Kirillov, Mintun, Ravi, Mao, Rolland, Gustafson,
  Xiao, Whitehead, Berg, Lo, et~al.]{kirillov2023segment}
Alexander Kirillov, Eric Mintun, Nikhila Ravi, Hanzi Mao, Chloe Rolland, Laura
  Gustafson, Tete Xiao, Spencer Whitehead, Alexander~C Berg, Wan-Yen Lo, et~al.
\newblock Segment anything.
\newblock \emph{arXiv preprint arXiv:2304.02643}, 2023.

\bibitem[Laskin et~al.(2020)Laskin, Lee, Stooke, Pinto, Abbeel, and
  Srinivas]{laskin2020reinforcement}
Misha Laskin, Kimin Lee, Adam Stooke, Lerrel Pinto, Pieter Abbeel, and Aravind
  Srinivas.
\newblock Reinforcement learning with augmented data.
\newblock \emph{Advances in neural information processing systems},
  33:\penalty0 19884--19895, 2020.

\bibitem[Lee et~al.(2020)Lee, Hwangbo, Wellhausen, Koltun, and
  Hutter]{lee2020learning}
Joonho Lee, Jemin Hwangbo, Lorenz Wellhausen, Vladlen Koltun, and Marco Hutter.
\newblock Learning quadrupedal locomotion over challenging terrain.
\newblock \emph{Science robotics}, 5\penalty0 (47):\penalty0 eabc5986, 2020.

\bibitem[Lee et~al.(2019)Lee, Lee, Shin, and Lee]{lee2019network}
Kimin Lee, Kibok Lee, Jinwoo Shin, and Honglak Lee.
\newblock Network randomization: A simple technique for generalization in deep
  reinforcement learning.
\newblock \emph{International Conference on Learning Representations}, 2019.

\bibitem[Li et~al.(2024)Li, Lyu, Ma, Wang, Yang, Li, and
  Li]{li2024normalization}
Lu Li, Jiafei Lyu, Guozheng Ma, Zilin Wang, Zhenjie Yang, Xiu Li, and Zhiheng
  Li.
\newblock Normalization enhances generalization in visual reinforcement
  learning.
\newblock In \emph{Proceedings of the 23rd International Conference on
  Autonomous Agents and Multiagent Systems}, pages 1137--1146, 2024.

\bibitem[Liu et~al.(2024)Liu, Jianye, Ma, and Xia]{liuunlock}
Jiashun Liu, HAO Jianye, Yi Ma, and Shuyin Xia.
\newblock Unlock the cognitive generalization of deep reinforcement learning
  via granular ball representation.
\newblock In \emph{International Conference on Machine Learning}. PMLR, 2024.

\bibitem[Liu et~al.(2023)Liu, Zhou, Yang, and Wang]{liu2023robust}
Qiyuan Liu, Qi Zhou, Rui Yang, and Jie Wang.
\newblock Robust representation learning by clustering with bisimulation
  metrics for visual reinforcement learning with distractions.
\newblock In \emph{Proceedings of the AAAI Conference on Artificial
  Intelligence}, pages 8843--8851, 2023.

\bibitem[Machado et~al.(2018)Machado, Bellemare, Talvitie, Veness, Hausknecht,
  and Bowling]{machado2018revisiting}
Marlos~C Machado, Marc~G Bellemare, Erik Talvitie, Joel Veness, Matthew
  Hausknecht, and Michael Bowling.
\newblock Revisiting the arcade learning environment: Evaluation protocols and
  open problems for general agents.
\newblock \emph{Journal of Artificial Intelligence Research}, 61:\penalty0
  523--562, 2018.

\bibitem[Makoviychuk et~al.(2021)Makoviychuk, Wawrzyniak, Guo, Lu, Storey,
  Macklin, Hoeller, Rudin, Allshire, Handa, et~al.]{makoviychuk2isaac}
Viktor Makoviychuk, Lukasz Wawrzyniak, Yunrong Guo, Michelle Lu, Kier Storey,
  Miles Macklin, David Hoeller, Nikita Rudin, Arthur Allshire, Ankur Handa,
  et~al.
\newblock Isaac gym: High performance gpu based physics simulation for robot
  learning.
\newblock In \emph{Thirty-fifth Conference on Neural Information Processing
  Systems Datasets and Benchmarks Track}, 2021.

\bibitem[Packer et~al.(2018)Packer, Gao, Kos, Kr{\"a}henb{\"u}hl, Koltun, and
  Song]{packer2018assessing}
Charles Packer, Katelyn Gao, Jernej Kos, Philipp Kr{\"a}henb{\"u}hl, Vladlen
  Koltun, and Dawn Song.
\newblock Assessing generalization in deep reinforcement learning.
\newblock \emph{arXiv preprint arXiv:1810.12282}, 2018.

\bibitem[Pinto et~al.(2017)Pinto, Andrychowicz, Welinder, Zaremba, and
  Abbeel]{pinto2017asymmetric}
Lerrel Pinto, Marcin Andrychowicz, Peter Welinder, Wojciech Zaremba, and Pieter
  Abbeel.
\newblock Asymmetric actor critic for image-based robot learning.
\newblock \emph{arXiv preprint arXiv:1710.06542}, 2017.

\bibitem[Raileanu and Rockt{\"a}schel(2020)]{raileanu2020ride}
Roberta Raileanu and Tim Rockt{\"a}schel.
\newblock Ride: Rewarding impact-driven exploration for procedurally-generated
  environments.
\newblock \emph{International Conference on Learning Representations}, 2020.

\bibitem[Rajeswaran et~al.(2017)Rajeswaran, Lowrey, Todorov, and
  Kakade]{rajeswaran2017towards}
Aravind Rajeswaran, Kendall Lowrey, Emanuel~V Todorov, and Sham~M Kakade.
\newblock Towards generalization and simplicity in continuous control.
\newblock \emph{Advances in Neural Information Processing Systems}, 30, 2017.

\bibitem[Ravi et~al.(2024)Ravi, Gabeur, Hu, Hu, Ryali, Ma, Khedr, R{\"a}dle,
  Rolland, Gustafson, et~al.]{ravi2024sam}
Nikhila Ravi, Valentin Gabeur, Yuan-Ting Hu, Ronghang Hu, Chaitanya Ryali,
  Tengyu Ma, Haitham Khedr, Roman R{\"a}dle, Chloe Rolland, Laura Gustafson,
  et~al.
\newblock Sam 2: Segment anything in images and videos.
\newblock \emph{arXiv preprint arXiv:2408.00714}, 2024.

\bibitem[Salter et~al.(2021)Salter, Rao, Wulfmeier, Hadsell, and
  Posner]{salter2021attention}
Sasha Salter, Dushyant Rao, Markus Wulfmeier, Raia Hadsell, and Ingmar Posner.
\newblock Attention-privileged reinforcement learning.
\newblock In \emph{Conference on Robot Learning}, pages 394--408. PMLR, 2021.

\bibitem[Schwarzer et~al.(2020)Schwarzer, Anand, Goel, Hjelm, Courville, and
  Bachman]{schwarzer2020data}
Max Schwarzer, Ankesh Anand, Rishab Goel, R~Devon Hjelm, Aaron Courville, and
  Philip Bachman.
\newblock Data-efficient reinforcement learning with self-predictive
  representations.
\newblock \emph{International Conference on Learning Representations}, 2020.

\bibitem[Srivastava et~al.(2014)Srivastava, Hinton, Krizhevsky, Sutskever, and
  Salakhutdinov]{srivastava2014dropout}
Nitish Srivastava, Geoffrey Hinton, Alex Krizhevsky, Ilya Sutskever, and Ruslan
  Salakhutdinov.
\newblock Dropout: a simple way to prevent neural networks from overfitting.
\newblock \emph{The journal of machine learning research}, 15\penalty0
  (1):\penalty0 1929--1958, 2014.

\bibitem[Stone et~al.(2021)Stone, Ramirez, Konolige, and
  Jonschkowski]{stone2021distracting}
Austin Stone, Oscar Ramirez, Kurt Konolige, and Rico Jonschkowski.
\newblock The distracting control suite--a challenging benchmark for
  reinforcement learning from pixels.
\newblock \emph{arXiv preprint arXiv:2101.02722}, 2021.

\bibitem[Vaswani et~al.(2017)Vaswani, Shazeer, Parmar, Uszkoreit, Jones, Gomez,
  Kaiser, and Polosukhin]{vaswani2017attention}
Ashish Vaswani, Noam Shazeer, Niki Parmar, Jakob Uszkoreit, Llion Jones,
  Aidan~N Gomez, {\L}ukasz Kaiser, and Illia Polosukhin.
\newblock Attention is all you need.
\newblock \emph{Advances in neural information processing systems}, 30, 2017.

\bibitem[Wang et~al.(2022)Wang, Luo, Ross, and Li]{wang2022vrl3}
Che Wang, Xufang Luo, Keith Ross, and Dongsheng Li.
\newblock Vrl3: A data-driven framework for visual deep reinforcement learning.
\newblock In \emph{Conference on Neural Information Processing Systems}, 2022.

\bibitem[Wang et~al.(2019)Wang, Zheng, Xiong, and
  Socher]{wang2019generalization}
Huan Wang, Stephan Zheng, Caiming Xiong, and Richard Socher.
\newblock On the generalization gap in reparameterizable reinforcement
  learning.
\newblock In \emph{International Conference on Machine Learning}, pages
  6648--6658. PMLR, 2019.

\bibitem[Wang et~al.(2021)Wang, Lian, and Yu]{wang2021unsupervised}
Xudong Wang, Long Lian, and Stella~X Yu.
\newblock Unsupervised visual attention and invariance for reinforcement
  learning.
\newblock In \emph{Proceedings of the IEEE/CVF Conference on Computer Vision
  and Pattern Recognition}, pages 6677--6687, 2021.

\bibitem[Yang et~al.(2024)Yang, Ze, and Xu]{yang2024movie}
Sizhe Yang, Yanjie Ze, and Huazhe Xu.
\newblock Movie: Visual model-based policy adaptation for view generalization.
\newblock \emph{Advances in Neural Information Processing Systems}, 36, 2024.

\bibitem[Yarats et~al.(2021{\natexlab{a}})Yarats, Fergus, Lazaric, and
  Pinto]{yarats2021mastering}
Denis Yarats, Rob Fergus, Alessandro Lazaric, and Lerrel Pinto.
\newblock Mastering visual continuous control: Improved data-augmented
  reinforcement learning.
\newblock In \emph{Deep RL Workshop NeurIPS 2021}, 2021{\natexlab{a}}.

\bibitem[Yarats et~al.(2021{\natexlab{b}})Yarats, Kostrikov, and
  Fergus]{yarats2021image}
Denis Yarats, Ilya Kostrikov, and Rob Fergus.
\newblock Image augmentation is all you need: Regularizing deep reinforcement
  learning from pixels.
\newblock In \emph{International Conference on Learning Representations},
  2021{\natexlab{b}}.

\bibitem[Yarats et~al.(2021{\natexlab{c}})Yarats, Zhang, Kostrikov, Amos,
  Pineau, and Fergus]{yarats2021improving}
Denis Yarats, Amy Zhang, Ilya Kostrikov, Brandon Amos, Joelle Pineau, and Rob
  Fergus.
\newblock Improving sample efficiency in model-free reinforcement learning from
  images.
\newblock In \emph{Proceedings of the AAAI Conference on Artificial
  Intelligence}, pages 10674--10681, 2021{\natexlab{c}}.

\bibitem[Yu et~al.(2021)Yu, Jain, Escontrela, Iscen, Xu, Coumans, Ha, Tan, and
  Zhang]{yu2021visual}
Wenhao Yu, Deepali Jain, Alejandro Escontrela, Atil Iscen, Peng Xu, Erwin
  Coumans, Sehoon Ha, Jie Tan, and Tingnan Zhang.
\newblock Visual-locomotion: Learning to walk on complex terrains with vision.
\newblock In \emph{5th Annual Conference on Robot Learning}, 2021.

\bibitem[Yuan et~al.(2022{\natexlab{a}})Yuan, Ma, Mu, Xia, Yuan, Wang, Luo, and
  Xu]{yuan2022don}
Z Yuan, G Ma, Y Mu, B Xia, B Yuan, X Wang, P Luo, and H Xu.
\newblock Don’t touch what matters: Task-aware lipschitz data augmentationfor
  visual reinforcement learning.
\newblock In \emph{Proceedings of the Thirty-First International Joint
  Conference on Artificial Intelligence, Vienna, 23-29 July 2022}.
  International Joint Conferences on Artificial Intelligence.,
  2022{\natexlab{a}}.

\bibitem[Yuan et~al.(2022{\natexlab{b}})Yuan, Xue, Yuan, Wang, Wu, Gao, and
  Xu]{yuan2022pretrained}
Zhecheng Yuan, Zhengrong Xue, Bo Yuan, Xueqian Wang, Yi Wu, Yang Gao, and
  Huazhe Xu.
\newblock Pre-trained image encoder for generalizable visual reinforcement
  learning.
\newblock In \emph{Advances in Neural Information Processing Systems},
  2022{\natexlab{b}}.

\bibitem[Yuan et~al.(2022{\natexlab{c}})Yuan, Xue, Yuan, Wang, Wu, Gao, and
  Xu]{yuanpre}
Zhecheng Yuan, Zhengrong Xue, Bo Yuan, Xueqian Wang, Yi Wu, Yang Gao, and
  Huazhe Xu.
\newblock Pre-trained image encoder for generalizable visual reinforcement
  learning.
\newblock In \emph{Advances in Neural Information Processing Systems},
  2022{\natexlab{c}}.

\bibitem[Zhang et~al.(2018{\natexlab{a}})Zhang, Ballas, and
  Pineau]{zhang2018dissection}
Amy Zhang, Nicolas Ballas, and Joelle Pineau.
\newblock A dissection of overfitting and generalization in continuous
  reinforcement learning.
\newblock \emph{arXiv preprint arXiv:1806.07937}, 2018{\natexlab{a}}.

\bibitem[Zhang et~al.(2020)Zhang, McAllister, Calandra, Gal, and
  Levine]{zhang2020learning}
Amy Zhang, Rowan McAllister, Roberto Calandra, Yarin Gal, and Sergey Levine.
\newblock Learning invariant representations for reinforcement learning without
  reconstruction.
\newblock \emph{arXiv preprint arXiv:2006.10742}, 2020.

\bibitem[Zhang et~al.(2018{\natexlab{b}})Zhang, Vinyals, Munos, and
  Bengio]{zhang2018study}
Chiyuan Zhang, Oriol Vinyals, Remi Munos, and Samy Bengio.
\newblock A study on overfitting in deep reinforcement learning.
\newblock \emph{arXiv preprint arXiv:1804.06893}, 2018{\natexlab{b}}.

\bibitem[Zhao et~al.(2023)Zhao, Kumar, Levine, and Finn]{zhao2023learning}
Tony~Z Zhao, Vikash Kumar, Sergey Levine, and Chelsea Finn.
\newblock Learning fine-grained bimanual manipulation with low-cost hardware.
\newblock \emph{arXiv preprint arXiv:2304.13705}, 2023.

\bibitem[Zhou et~al.(2020)Zhou, Kalra, and Kr{\"a}henb{\"u}hl]{zhou2020domain}
Brady Zhou, Nimit Kalra, and Philipp Kr{\"a}henb{\"u}hl.
\newblock Domain adaptation through task distillation.
\newblock In \emph{Computer Vision--ECCV 2020: 16th European Conference,
  Glasgow, UK, August 23--28, 2020, Proceedings, Part XXVI 16}, pages 664--680.
  Springer, 2020.

\end{thebibliography}
}

\clearpage
\setcounter{page}{1}
\maketitlesupplementary

\section{Implementation Details}\label{sec:supp_a}

In this section, we first provide the implementation details for \ourmethod, and then we provide the implementation details for a strong baseline SAM+E we utilized in our experiments.

\subsection{Implementations of \ourmethod}

\textbf{Algorithm details}. We present the pseudocode of \ourmethod~ in Algorithm~\ref{pseudoPSO}.
We first train the privileged expert given only the state information (i.e., the state $s$) using DrQv2~\citep{yarats2021mastering} (from line 5 to line 12). Note that any other standard RL algorithm can be applied here to replace DrQv2. In the second part of \ourmethod~ (from line 13 to line 24), we train the control-aware data augmentation module and privilege-guided policy distillation module \textbf{simultaneously}.
For the former module, we build our implementation on top of the convolutional block attention module (CBAM) which consists of a channel attention block and a spatial attention block.
The channel attention block comprises an averaging operation module (an average-pooling function followed by two MLP layers) and a maxing operation module (a max-pooling function followed by two MLP layers). A sigmoid activation function is applied on the end to produce a channel mask.
The spatial attention block comprises a channel pool function (to extract the maximum and average features across channels), convolutional layers (with a kernel size of 7, stride of 1, and padding of 3), and a batchnorm layer sequentially. More detailed network structures and parameter settings can be found in our open-source code.

\textbf{Hyperparamter settings}. In practice, we build our implementation on top of DrQv2~\citep{yarats2021mastering} due to its broad adoption in continuous control tasks~\citep{wang2022vrl3,yuanpre,haldar2023watch} as well as a clean and efficient codebase. Therefore, \ourmethod~ follows the hyperparameter settings from DrQ-v2, as listed in Tab.~\ref{tab:para_CACD}. Specifically, \ourmethod~introduces only one new hyperparameter $\lambda$ to control the sparsity of the attention mask. After conducting ablation studies on the sparsity loss in Supplementary~\ref{sec:supp_b}, we set $\lambda$ to 0.001 for DMC-GB/DOGB and 0.01 for RMDB. 
For data augmentation methods, we use random overlay with Places365 for DMC-GB/DOGB and random conv for RMDB.

\begin{algorithm}
\caption{\ourmethod}
\label{pseudoPSO}
\begin{algorithmic}[1]
\State $\pi_e$, $Q_1$, $Q_2$: randomly initialized policy network and two value network for privileged expert
\State $f_e(\cdot)$, $f_d(\cdot)$, $f_a(\cdot)$,$f_{ctl}(\cdot, \cdot)$, : randomly initialized encoder, decoder, attention network and control prediction network in control-aware augmentation module
\State $\pi$: random initialized student policy network
\State replay buffer $\mathcal{D}$, mini-batch size $b$
\For{Each interaction timestep $i$}
    \State Rollout $\pi_e$ for one timestep and add ($s_i$, $a_i$, $r_i$) to dataset $\mathcal{D} \leftarrow \mathcal{D} \cup (s_i, a_i, r_i)$
    \If {$|\mathcal{D}| \textgreater b$}
        \State Sample experiences $(\textbf{s}, \textbf{a}, \textbf{r}) \sim \mathcal{D}$
        \State Update $\pi_e$ 
        \State Update $Q_1$, $Q_2$ 
    \EndIf
\EndFor

\State Empty $\mathcal{D}=\emptyset$
\For{Each interaction timestep $t$}
    \State Choose $\pi_{roll}$ from $\pi_e$ or $\pi$
    \State Rollout $\pi_{roll}$ for one timestep and add ($o_t$, $s_t$) to dataset $\mathcal{D} \leftarrow \mathcal{D} \cup (o_t, s_t)$
    \If {$|\mathcal{D}| \textgreater b$}
        \State Sample experiences $(\textbf{o},\textbf{s}, \textbf{o}') \sim \mathcal{D}$
        \State Update control-aware data augmentation module according to Equation~\ref{equ:att_loss}
        \State Obtain control-aware augmented observation $\textbf{o}_{aug}$ according to Equation~\ref{equ:aug}
        \State Update $\pi$ according to Equation~\ref{equ:cloning_loss}
    \EndIf
\EndFor
\end{algorithmic}
\end{algorithm}

\begin{table}[h]
\caption{\ourmethod~ Hyperparamters.}
\label{tab:para_CACD}
\centering
\resizebox{0.9\columnwidth}{!}{
    \begin{tabular}{cll}
    \toprule
     & Hyperparameter & Value \\ 
    \midrule
    \multirow{8}*{\shortstack{Teacher}} 
    & Learning rate for all net & 1e-4 \\ 
    & Target update rate & 0.01 \\ 
    & Optimizer & Adam \\ 
    & Batch size & 256 \\ 
    & Discount factor & 0.99 \\ 
    & n-step & 1 for Walker walk $\&$ Walker stand, 3 for others \\ 
    & Update interval & 2 \\
    & Replay buffer size & 500k \\
    \midrule
    \multirow{10}*{Student}
    & Observation & 84 $\times$ 84 for DMC-GB/RMDB, 128 $\times$ 128 for DOGB \\ 
    & Learning rate for all net & 1e-4 \\ 
    & Optimizer & Adam \\
    & Batch size & 256 \\ 
    & Frame stack & 3 for DMC-GB, 1 for RMDB/DOGB \\ 
    & Update interval & 2 \\
    & Replay buffer size & 500k for DMC-GB/RMDB, 100k for DOGB\\
    & $\lambda$ & 0.001 for DMC-GB, 0.01 for RMDB/DOGB \\
    & $\beta$ & 0.5 \\
    & $\alpha$ in $\textit{random overlay}$ & linear schedule from 0.4 to 0.9 \\  
    \bottomrule
    \end{tabular}
}
\end{table} 

\begin{figure*}[t]
    \centering
    \includegraphics[width=1.0\textwidth]{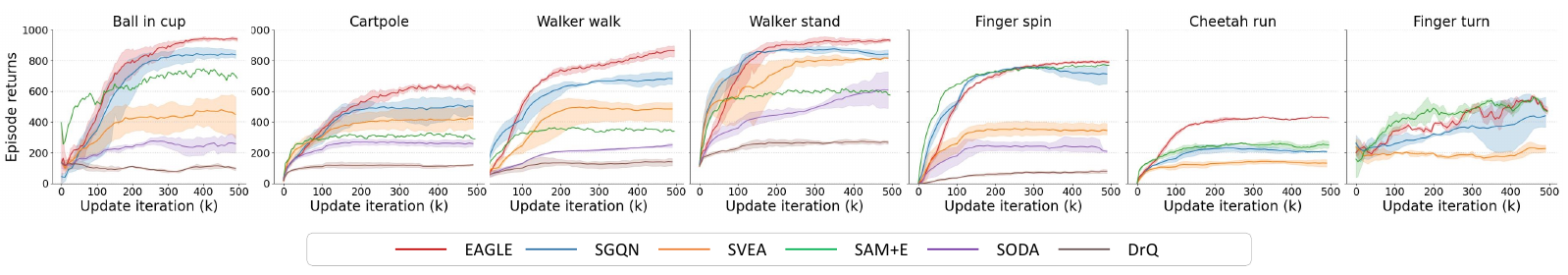}
    \caption{Generalization performance of all methods on DMC-GB video-hard.}
    \label{fig_dmcgb-main}
\end{figure*}

\begin{figure*}[t]
    \centering
    \includegraphics[width=1.0\textwidth]{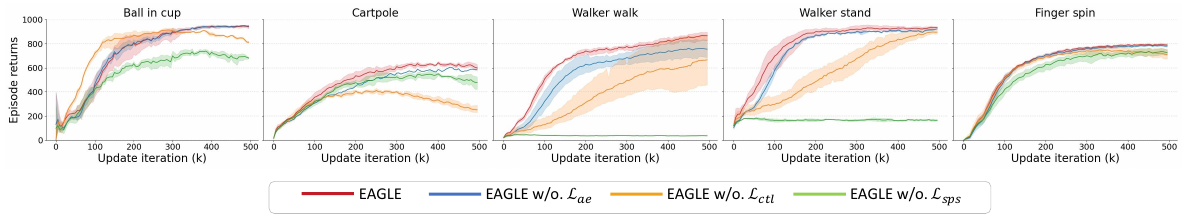}
    \caption{Ablation study on augmentation loss under DMC-GB video-hard.}
    \label{fig:aug_loss}
\end{figure*}

\subsection{Implementations of SAM+E}\label{sec:sam_e}
SAM~\cite{kirillov2023segment} is one of the most popular state-of-the-art~(SOTA) large vision models that is capable of getting the segmentation results automatically with proper point prompts over various domains. Therefore, we use SAM to construct a strong baseline SAM+E. Considering the mask inference time, we use the ViT-B SAM model checkpoint. To get proper background prompts, we first upload an example image on Segment-Anything website and randomly sample points until we get a clear segmentation result. Then, we use an Image Position Coordinate Tool to get these prompts' positions. After configuring the point prompts, we check the segmentation results again with Weights\&Biases.

\begin{figure}
    \centering
    \includegraphics[width=1.0\columnwidth]{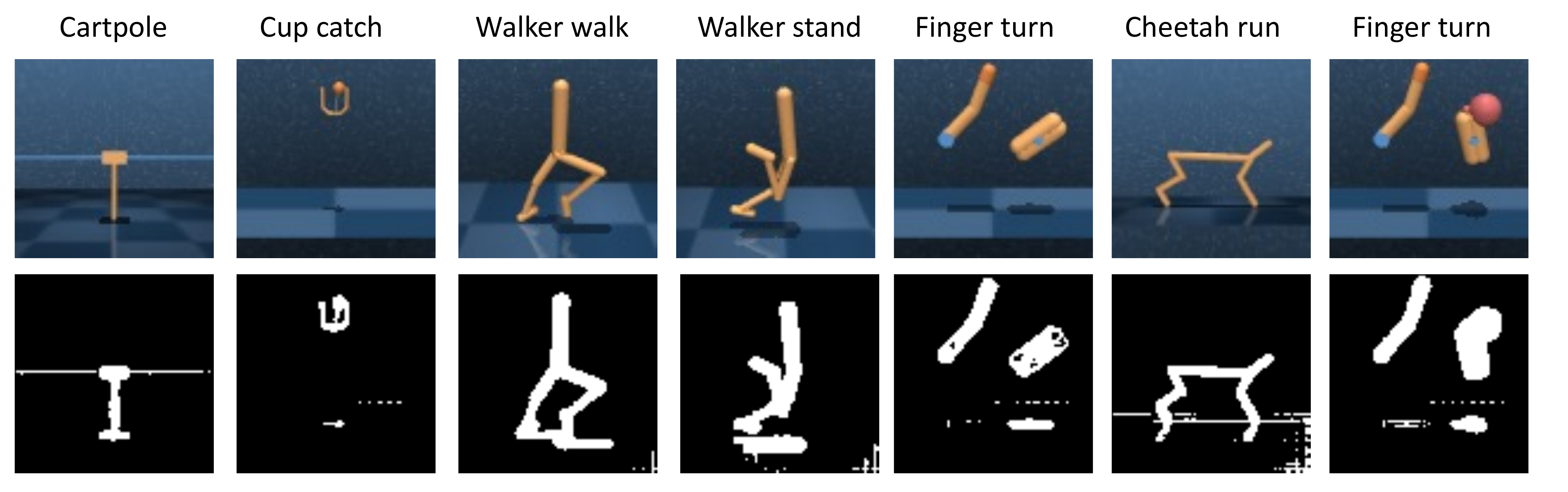}
    \caption{Segmentation results in DMC-GB with SAM.}
    \label{fig:sam-seg}
\end{figure}

The prompts we used in DMC-GB are (2, 80), (2, 57), (3, 7). The corresponding segmentation results we got with SAM are shown in Fig.~\ref{fig:sam-seg}. The prompts we used in RMDB are (2, 5), (2, 24), (2, 37), (2, 67). The corresponding segmentation results are shown in Fig.~\ref{fig:sam-seg-robot}.

\section{Additional results on DMC-GB}\label{sec:supp_b}
In this section, we first provide more ablation results on DMC-GB. Then we provide more visualization results with \ourmethod~in DMC-GB.

\subsection{Ablation study on the loss function}
The original manuscript presented ablation results on the loss function for the Walker-walk and Walker-stand tasks. In this section, we extend the ablation analysis to five tasks in DMC-GB, focusing on the loss terms $\mathcal{L}_{rec}$, $\mathcal{L}_{ctl}$, and $\mathcal{L}_{sps}$ in Equ.~\ref{equ:att_loss}. 

\textbf{Impact of $\mathcal{L}_{ctl}$}. Apparently, the control prediction loss $\mathcal{L}_{ctl}$ (represented by the orange line in Fig.~\ref{fig:aug_loss}) is crucial for the stability performance of \ourmethod. Absence of $\mathcal{L}_{ctl}$ may result in significant variance in policy generalization performance (as observed in Walker walk and Walker stand tasks) or lead to a performance decline during training (as shown in Ball in cup, Cartpole, and Finger spin tasks).
This is due to the sparsity constraint applied in our method.  This constraint may cause the augmentation module to distribute low values evenly across the entire dynamic foreground in \ourmethod~w/o. $\mathcal{L}_{ctl}$. Consequently, with a linear growing schedule in \textit{random overlay}, some control-related parts (e.g., the tip of the cartpole) may be disrupted by the augmentation process, thus influencing the overall stability and performance of the model.

\textbf{Impact of $\mathcal{L}_{sps}$}. The sparsity loss $\mathcal{L}_{sps}$ (represented by the green line in Fig.~\ref{fig:aug_loss}) is also crucial for generalization performance. In the absence of $\mathcal{L}_{sps}$, the model's generalization ability may converge to a lower level or even experience policy degradation. This occurs because, in the absence of the sparsity constraint, the generated mask may focus on a control-irrelevant background or even the entire observation space (in the Walker environment). Without fully utilizing the augmentation technique, this, in turn, leads to an apparent performance decline in video-hard settings.

\textbf{Impact of $\mathcal{L}_{ae}$}. The effect of $\mathcal{L}_{ae}$ (represented by the blue line in Fig.~\ref{fig:aug_loss}) is different across five tasks. Overall, by adding the auto-encoder loss, we could learn a stable and generalizable policy in various tasks.

\begin{figure}
    \centering
    \includegraphics[width=1.0\columnwidth]{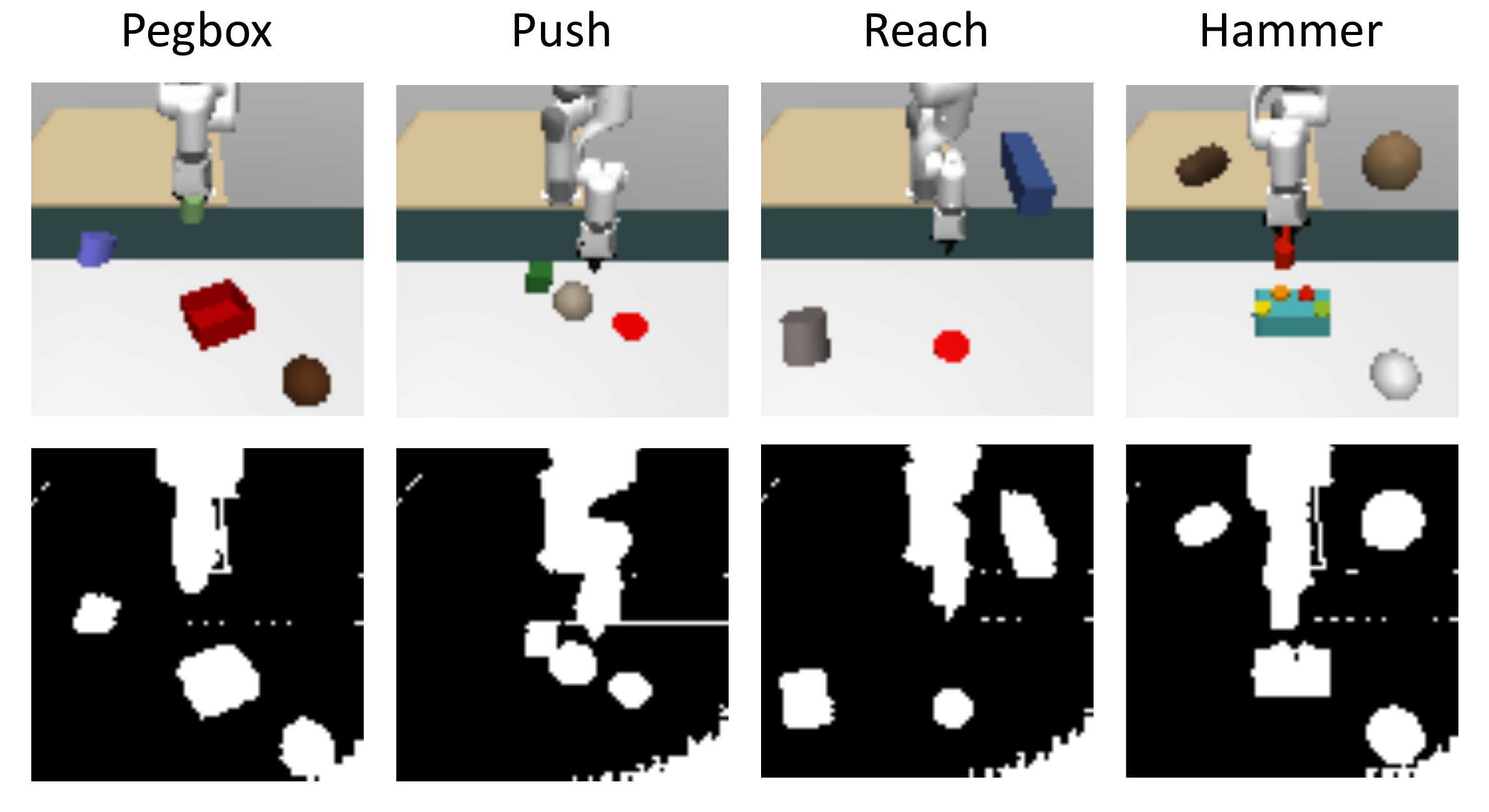}
    \caption{Segmentation results in RMDB with SAM.}
    \label{fig:sam-seg-robot}
\end{figure}

\subsection{Visualization of \ourmethod~ on DMC-GB}
\begin{figure}
    \centering
    \includegraphics[width=1.0\columnwidth]{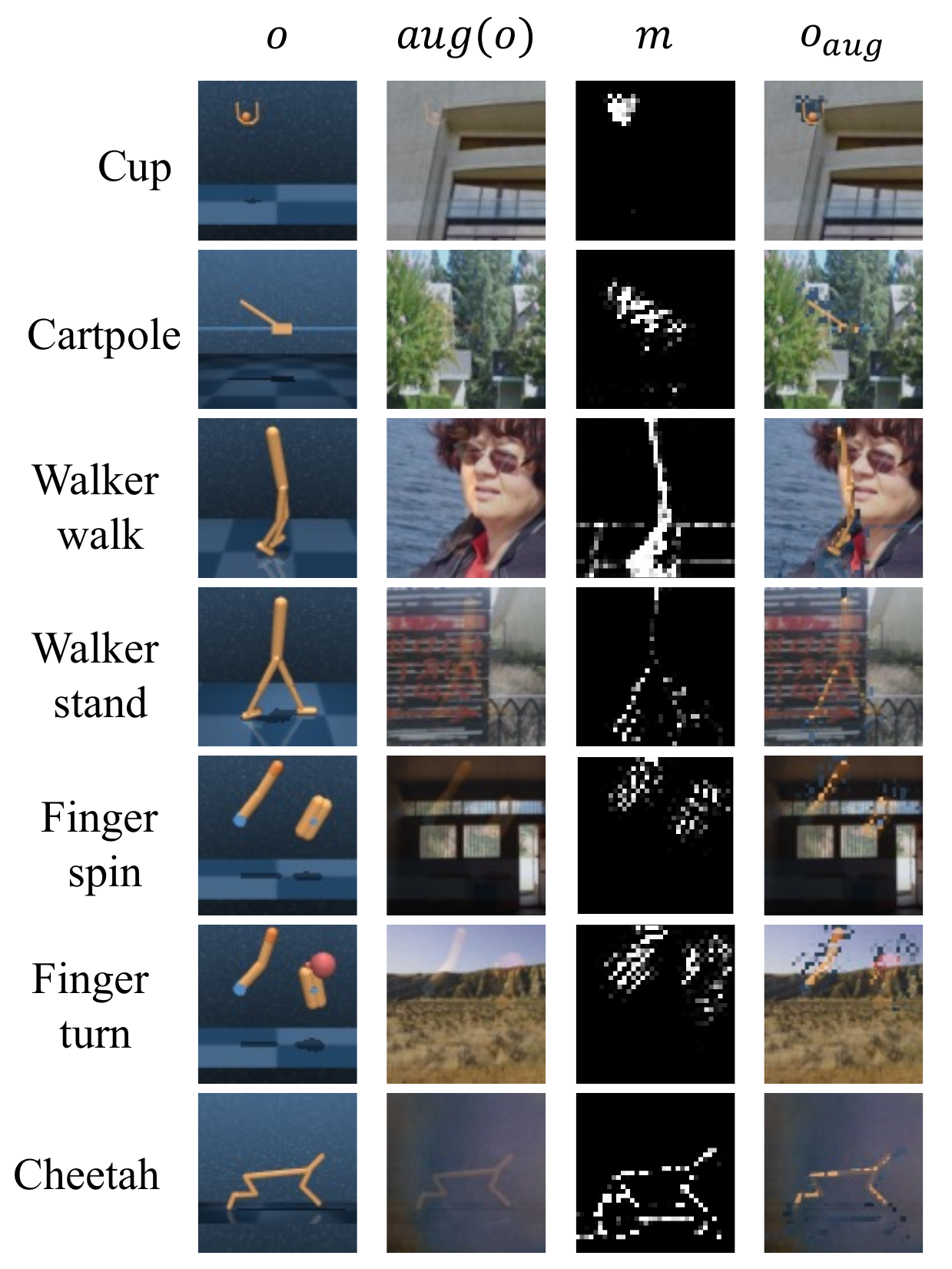}
    \caption{Visualization results of \ourmethod~in DMC-GB.}
    \label{fig:gemo-seg}
\end{figure}

In Fig.~\ref{fig:gemo-seg}, we visualize the original observation $o$, the vanilla augmented observation $aug(o)$ with \textit{random overlay} ($\alpha$=0.8), the control-aware mask $m$ and the control-aware augmented observation $o_{aug}$. We clearly observed that, with our control-aware mask, \ourmethod~augments only the control-irrelevant regions while preserving the control-related parts. For example, when the walker is standing (in Walker stand), we preserve the some region of the "leg" and "head" of the robot. While the walker is walking (in Walker walk), we mainly concentrate on the "leg" and the "upper-body". With this control-aware augmentation module, we preserve the most control-related features for decision-making while augmenting other areas to enhance generalization.

\section{Additional results on RMDB}\label{sec:supp_c}
\subsection{Detail description of RMDB}
Compared with the original Robot Manipulation benchmark, we made the following modifications in RMDB:
\begin{itemize}
    \item For each task, we introduced 1-3 distractors on the front desk and/or background floor. The color of these distractors varies in each testing environment but remains consistent during training. We selected this configuration to assess the policy's generalization ability when facing visual appearance changes of these distractors.
    \item We revised the reward function in the Push task. The original function only considered control magnitude and the distance between the object and its target. We introduced an additional reward term to incentivize the robot arm to approach the object, facilitating learning.
\end{itemize}

Despite the aforementioned modifications, we follow the original Robot Manipulation benchmark to validate the policy across five testing environments, each with different visual changes to the front desk, background wall, and operating objects. The training environment examples can be found in Fig.~\ref{fig:sam-seg-robot}. The testing environment examples can be found in Fig.~\ref{fig_all_env}.

\subsection{Visualization of \ourmethod~on RMDB}
\begin{figure}
    \centering
    \includegraphics[width=1.0\columnwidth]{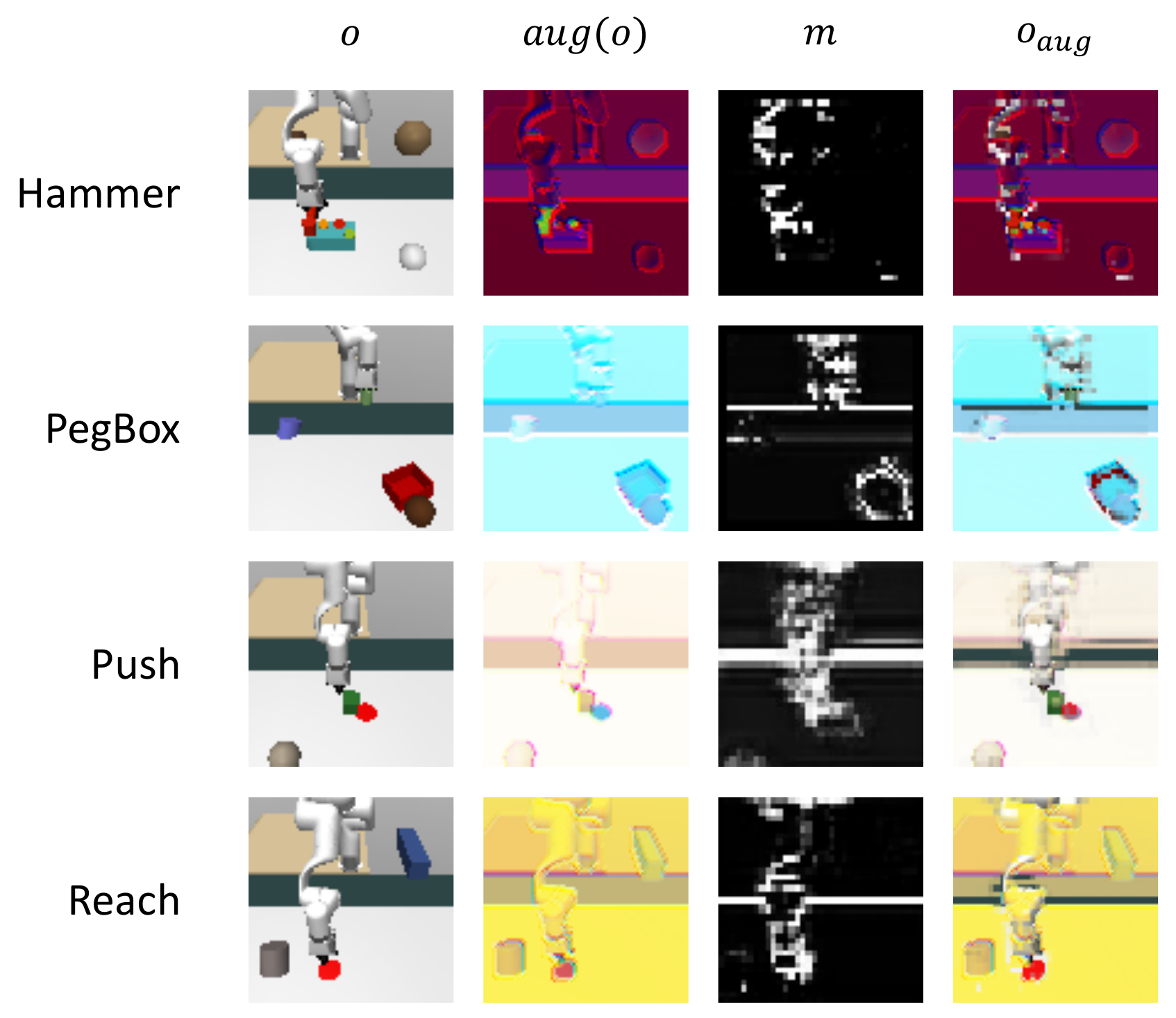}
    \caption{Visualization results of \ourmethod~in RMDB.}
    \label{fig:vis_robot}
\end{figure}

In Fig.~\ref{fig:vis_robot}, we demonstrate the original observation $o$, the vanilla augmented observation $aug(o)$ with \textit{random conv}, the control-aware mask $m$ and the control-aware augmented observation $o_{aug}$ in RMDB. More detailed analysis and the comparison with the mask obtained by SAM can be found in Section.~\ref{sec:compare} in the manuscript.

\section{Real world experiments}
\begin{figure}[hpt]
    \centering
    \subfloat[Left view]{\includegraphics[width=0.3\textwidth]{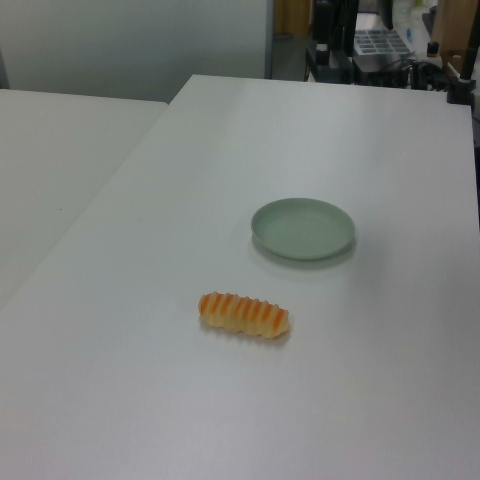}\label{fig:real-left}}
    \hfill
    \subfloat[Topdown view]{\includegraphics[width=0.3\textwidth]{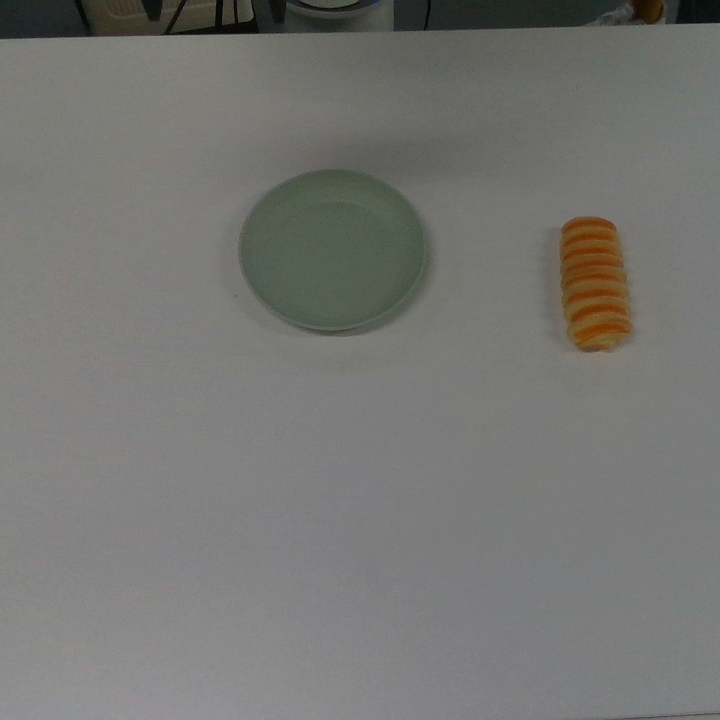}\label{fig:real-top}}
    \hfill
    \subfloat[right view]{\includegraphics[width=0.3\textwidth]{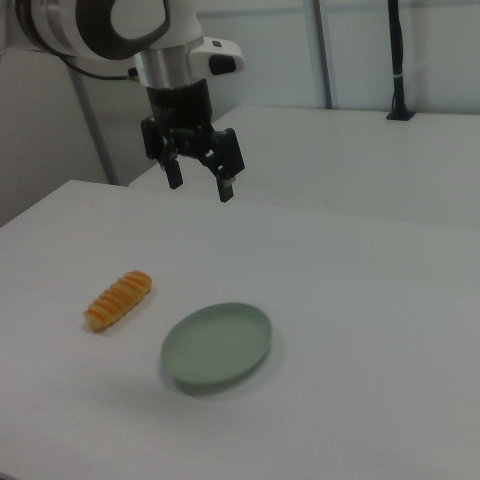}\label{fig:real-right}}
    \caption{Three camera views used in EAGLE.}
    \label{fig:camera-view}
\end{figure}
To validate \ourmethod~in real-world scenarios, we select a typical robot manipulation task of picking bread and placing it on a plate. We collect 200 trajectories using an expert policy and perform only offline distillation for safety considerations. As shown in Fig.~\ref{fig:camera-view}, the student visuomotor policy receives inputs from three cameras (top: cropped to $720\times720$; left and right: cropped to $480\times480$) and the joint positions of the Franka Arm. The output action is a continuous vector consisting of seven joint positions and one gripper value. All image inputs are resized to $256\times256$. The batch size is set to 16, and other parameters are the same as DOGB in Tab.~\ref{tab:para_CACD}. We use ACT~\cite{zhao2023learning} with ResNet50 as the backbone, and combine it with our control-aware augmentation module and Random Conv techniques. We compare the generalization performance of \ourmethod~and ACT by averaging results over 20 trials with varying bread positions. As shown in Tab.~\ref{tab:real-world}, \ourmethod~significantly improves generalization when facing with changing backgrounds and added distractors, whereas the baseline ACT lacks such generalization without proper data augmentation.

\begin{table}
\centering
\resizebox{0.95\columnwidth}{!}{
\begin{tabular}{lcccccc} 
\toprule
\textbf{Method} & \textbf{Training} & \textbf{Testing (Change Background)} & \textbf{Testing (Add Distractors)}\\
\midrule
\textbf{\ourmethod} & 0.85 & 0.75 & 0.60\\
\textbf{ACT~\cite{zhao2023learning}} & 0.80 & 0.00 & 0.00\\
\bottomrule
\end{tabular}
}
\caption{Real-world Experiments.}
\label{tab:real-world}
\end{table}

\end{document}